\documentclass[final, 12pt]{elsarticle}
\graphicspath{ {./img/} }
\usepackage{hyperref}
\usepackage{float}
\usepackage{verbatim} 
\usepackage{apalike}
\restylefloat{figure}
\restylefloat{table}

\usepackage{graphicx}
\usepackage{amssymb}
\usepackage{amsmath}
\usepackage{multirow}
\usepackage{comment}
\usepackage{tabularx}
\usepackage{ltablex}
\usepackage{xcolor}
\usepackage{pst-node}
\usepackage{tikz}
\usetikzlibrary{trees}
\usepackage{url}
\usepackage{booktabs}
\usepackage{pgfplots}
\usepackage{subcaption}

\journal{Expert Systems with Applications}

\pgfplotsset{compat=1.14}

\biboptions{authoryear}

\begin{document}
\begin{frontmatter}

\title{A machine learning approach for forecasting hierarchical time series}

\author[1]{Paolo Mancuso}
\ead{paolo.mancuso@uniroma2.it}

\author[2]{Veronica Piccialli}
\ead{veronica.piccialli@uniroma2.it}

\author[2]{Antonio M. Sudoso}
\ead{antonio.maria.sudoso@uniroma2.it}

\address[1]{Department of Industrial Engineering, University of Rome Tor Vergata, Italy}

\address[2]{Department of Civil Engineering and Computer Science Engineering, University of Rome Tor Vergata, Italy}

\begin{abstract}
In this paper, we propose a machine learning approach for forecasting hierarchical time series. When dealing with hierarchical time series, apart from generating accurate forecasts, one needs to select a suitable method for producing reconciled forecasts. Forecast reconciliation is the process of adjusting forecasts to make them coherent across the hierarchy. In literature, coherence is often enforced by using a post-processing technique on the base forecasts produced by suitable time series forecasting methods. On the contrary, our idea is to use a deep neural network to directly produce accurate and reconciled forecasts. We exploit the ability of a deep neural network to extract information capturing the structure of the hierarchy. We impose the reconciliation at training time by minimizing a customized loss function. In many practical applications, besides time series data, hierarchical time series include explanatory variables that are beneficial for increasing the forecasting accuracy. Exploiting this further information, our approach links the relationship between time series features extracted at any level of the hierarchy and the explanatory variables into an end-to-end neural network providing accurate and reconciled point forecasts. 
The effectiveness of the approach is validated on three real-world datasets, where our method outperforms state-of-the-art competitors in hierarchical forecasting.


\end{abstract}

\begin{keyword}
Hierarchical Time Series \sep Forecast \sep Machine Learning \sep Deep Neural Network
\end{keyword}

\end{frontmatter}

\section{Introduction}

A hierarchical time series is a collection of time series organized in a hierarchical structure that can be aggregated at different levels \citep{hyndman_forecasting}. As an example, Stock Keeping Unit (SKU) sales aggregate up to product subcategory sales, which further aggregate to product categories \citep{franses2011combining}.
Hierarchical forecasting is a very important application of expert systems for decision-making \citep{huber2017cluster}.
In order to support decision-making at different levels of the hierarchy, a challenging task is the generation of coherent forecasts. Forecasts of the individual series are coherent when they sum up in a proper way across the levels preserving the hierarchical structure. 

Coherence can be required either at the cross-sectional level 
or at the temporal level. For example, at the cross-sectional level, forecasts of regional sales should sum up to give forecasts of state sales, which should, in turn, sum up to give forecasts for the national sales. For temporal coherence instead, forecasts at the day level must sum up coherently at the week level, then at the month level, and so on. Recently, hierarchical time series attracted attention, see \cite{hollyman2021} and references therein. Usually, the two types of coherence are pursued with different and dedicated approaches, apart from some recent papers \citep{kourentzes2019cross, di2020cross, Spiliotis2020}. In this paper, we focus on cross-sectional coherence.
In the literature, two lines of research among others are pursued: top-down and bottom-up approaches. Top-down approaches involve forecasting first the top-level series and then disaggregating by means of historical \citep{gross1990disaggregation} or forecasted proportion \citep{athanasopoulos2009hierarchical} to get forecasts for the lower-level series. On the other hand, the bottom-up approach produces first forecasts for the bottom-level time series and then aggregates them to get the forecasts for the higher-level time series. Both classes of methods have their advantages since top-down approaches perform well when the top-level series is easy to forecast, whereas the bottom-up method accurately identifies the pattern of each series without loss of information. However, the bottom-up approach ignores correlations among the series, possibly leading to aggregate forecasts worse than the ones produced by top-down approaches \citep{shlifer}. In general, a bottom-up approach should be preferable whenever the forecasts are employed to support decisions that are mainly related to the bottom rather than the top of the hierarchy, whereas a top-down approach performs better when the bottom-level series are too noisy \citep{dunn}. 
The objective to reconcile forecasts at all levels of the hierarchy has lead researchers to investigate the impact that the association between bottom-level series produces on the aggregation \citep{Nenova2016}. Analytical approaches to the forecast reconciliation problem have been proposed by \cite{hyndman2011optimal} and by \cite{wickramasuriya2019optimal}. These methods not only ensure that forecasts are coherent but also lead to improvements in forecast accuracy. However, a shortcoming of these methods is the need for two stages, with forecasts first produced independently for each series in the hierarchy, and then optimally combined to satisfy the aggregation constraint. Therefore, the reconciliation is the result of post-processing on the base forecasts.  In \cite{hollyman2021}, all the above-mentioned methods are reconsidered within the framework of forecast combinations, showing that they can all be re-interpreted as particular examples of forecast combination where the coherence constraint is enforced with different strategies. The authors also show that  combining forecasts at the bottom level of the hierarchy can be exploited to improve the accuracy of the higher levels. \par

In recent years, machine learning models, especially based on neural networks, have emerged in the literature as an alternative to statistical methods for forecasting non-hierarchical time series. Indeed, many papers define new machine learning algorithms (see for example  \cite{bontempi2012machine,liu2020dstp,bandara2020forecasting,carta2021multi,ye2021implementing}), proposing innovative forecasting strategies that aim at improving the accuracy of time series predictions. Drawing inspiration from this line of research, we propose a machine learning approach for forecasting hierarchical time series. Using machine learning in hierarchical time series has also been considered recently in \cite{spiliotis2020hierarchical}. The authors propose a bottom-up method where the forecasts of the series of the bottom level are produced by a machine learning model (Random Forest and XGBoost), taking as input the base forecasts of all the series of the hierarchy. The reconciliation is then obtained by summing up the bottom-level forecasts.\par
Rather than formulating the reconciliation problem as a post-processing technique or just forecasting the bottom-level time series, our idea is to define a method that can automatically extract at any level of the hierarchy all the relevant information, keeping into account during the training also the reconciliation. Furthermore, our approach is able to easily incorporate at any level the information provided by the explanatory variables.

Forecasting models for time series with explanatory variables aim to predict correlated data taking into account additional information, known as exogenous variables. It is well known that incorporating explanatory variables in time series models helps to improve the forecast accuracy (see \cite{maccaira2018time} for a systematic literature review), thus in this paper we focus on these types of time series in the context of hierarchical forecasting. Our idea is to combine the explanatory variables with time series features defining the structure of the hierarchy to enhance the reconciliation and forecasting process. The main instrument we use to extract time series features is a Deep Neural Network (DNN). DNNs are designed to learn hierarchical representations of data \citep{lecun2015deep}. Thanks to the ability to extract meaningful features from data, Convolutional Neural Networks (CNNs) have been successful in time series forecasting and classification producing state-of-the-art results \citep{fawaz2019deep} but they have not been used in hierarchical time series forecasting. Our intuition is that extracting information at any level of the hierarchy through a CNN can be used to discover the structure of the series below.

\par
Hierarchical forecasting is relevant in many applications, such as energy and tourism, and it is common in the retail industry where the SKU demand can be grouped at different levels. Therefore, we prove the effectiveness of our method using three public datasets coming from real-world applications. The first one considers five years of sales data of an Italian grocery store, has three levels and noisy bottom-level series. This dataset has been made public by the authors (see \cite{mancuso2021}). The second one has two levels and comes from electricity demand data in Switzerland \citep{nespoli2020hierarchical}; it has quite regular bottom-level series, whereas the third one with four levels is extracted from the Walmart data used in the M5 forecasting competition. On all these datasets, our method increases the forecasting accuracy of the hierarchy outperforming state-of-the-art approaches, as confirmed by deep statistical analysis.

\par
However, our methodology for forecasting hierarchical time series shares the same limitations of the machine learning approaches for forecasting non-hierarchical time series: it is not viable for time series with a too-small number of historical observations (i.e. a few years of observations for daily time series are needed).\par Summarizing, the main contributions of the paper are:
\begin{enumerate}
    \item  For the first time, we introduce the use of machine learning in the forecasting of hierarchical time series, defining a methodology that can be used at any level of the hierarchy to generate accurate and coherent forecasts for the lower levels.
    \item Our method uses a deep neural network that is able at once to automatically extract the relevant features of the hierarchy while forcing the reconciliation and easily exploiting the exogenous variables at any level of the hierarchy.
    \item We consider three real-world datasets, and we perform comparisons with state-of-the-art methods in hierarchical forecasting. Furthermore, a deep statistical analysis assesses the superiority of our approach in comparison to standard methods. \item We share with the research community a new challenging dataset for hierarchical forecasting coming from the sales data of an Italian grocery store.
\end{enumerate}

The rest of the paper is organized as follows. Section 2 discusses the concept of hierarchical time series and the methods of hierarchical forecasting. Section 3 contains the detail of the proposed machine learning algorithm. Section 4 describes the basic forecasting methods employed in the hierarchical models and the experimental setup. Section 5 discusses the datasets and the numerical experiments conducted to evaluate the proposed method. Finally, Section 6 concludes the paper.

\section{Hierarchical Time Series}\label{sec:hier}

In a general hierarchical structure with $K>0$ levels, level 0 is defined as the completely aggregated series. Each level from 1 to $K-2$ denotes a further disaggregation down to level $K-1$ containing the most disaggregated time series. In a hierarchical time series, the observations at higher levels can be obtained by summing up the series below. Let $\boldsymbol{y_t^k}\in\Re^{m_k}$ be the vector of all observations at level $k = 1, \dots, K-1$ and $t = 1, \dots, T$, where $m_k$ is the number of series at level $k$ and $M = \sum_{k=0}^{K-1} m_k$ is the total number of series in the hierarchy. Then we define the vector of all observations of the hierarchy:
\[
\boldsymbol{y_t} =\begin{pmatrix}
y_t^0\cr \boldsymbol{y_t^1}\cr \vdots\cr \boldsymbol{y_t^{K-1}}\end{pmatrix},
\]
where $y_t^0$ is the observation of the series at the top and the vector $\boldsymbol{y_t^{K-1}}$ contains the observations of the series at the bottom of the hierarchy. 
The structure of the hierarchy is determined by the summing matrix $\boldsymbol{S}$ that defines the aggregation constraints:
\[
\boldsymbol{y_t} = \boldsymbol{Sy_t^{K-1}}.
\]
The summing matrix $\boldsymbol{S}$ is a matrix having entries belonging to $\{0,1\}$ of size $M \times m_{K-1}$. 

Given observations at time $t = 1, ..., T$ and the forecasting horizon $h$, the aim is to forecast each series at each level at time $t = T + 1, . . . , T + h$. The current methods of forecasting hierarchical time series are top-down, bottom-up, middle-out, and optimal reconciliation \citep{hyndman_forecasting, hollyman2021}. The main objective of such approaches is to ensure that forecasts are coherent across the levels of the hierarchy. Regardless of the methods used to forecast the time
series for the different levels of the hierarchy, the individual forecasts must be reconciled to be
useful for any subsequent decision making. Forecast reconciliation is the process of adjusting forecasts to make them coherent. By definition, a forecast is coherent if it satisfies the aggregation constraints defined by the summing matrix.

\subsection{Bottom-up Approach}

The bottom-up approach focuses on producing the $h$-step-ahead base forecasts for each series at the lowest level $\boldsymbol{\hat{y}_h^{K-1}}$ and aggregating them to the upper levels of the hierarchy according to the summing matrix. It can be represented as follows:
\[
\boldsymbol{\Tilde{y}_h} = \boldsymbol{S\hat{y}_h^{K-1}},
\]
where $\boldsymbol{\Tilde{y}_h}$ is the vector of coherent $h$-step-ahead forecasts for all series of the hierarchy. An advantage of this approach is that we directly forecast the series at the bottom level, and no information gets lost due to the aggregation. On the other hand, bottom-level series can be noisy and more challenging to model and forecast. This approach also has the disadvantage of having many time series to forecast if there are many series at the lowest level.
\subsection{Top-down Approaches}

Top-down approaches first involve generating the base forecasts for the total series and then disaggregating these downwards to get coherent forecasts for each series of the hierarchy.
The disaggregation of the top-level forecasts is usually achieved by using the proportions $\boldsymbol{p} = (p_1,...,p_{m_{K-1}})^\mathsf{T}$, which represent the relative contribution of the bottom-level series to the top-level aggregate. The two most commonly used top-down approaches are the Average Historical Proportions (AHP) and the Proportions of the Historical Averages (PHA). In the case of the AHP, the proportions are calculated as follows: 
\[
p_i = \frac{1}{T}\sum_{t=1}^{T}\frac{y_{t, i}^{K-1}}{y_t^0}, \quad i=1, \dots, m_{K-1}.
\]
In the PHA approach, the proportions are obtained in the following manner:
\[
p_i = \frac{\sum_{t=1}^{T}\frac{y_{t, i}^{K-1}}{T}}{\sum_{t=1}^{T}\frac{y_{t}^{0}}{T}}, \quad i=1, \dots, m_{K-1}.
\]
For these two methods, once the bottom-level $h$-step-ahead forecasts have been generated, these are aggregated to produce coherent forecasts for the rest of the series of the hierarchy by using the summing matrix. Given the vector of proportions $\boldsymbol{p}$, top-down approaches can be represented as: 
\[
\boldsymbol{\Tilde{y}_h = S p }\hat{y}_h^{0}.
\]
Top-down approaches based on historical proportions usually produce less accurate forecasts at lower levels of the hierarchy than bottom-up approaches because they don't take into account that these proportions may change over time. To address this issue, instead of using the static proportions as in AHP and PHA, \cite{athanasopoulos2009hierarchical} propose the Forecasted Proportion (FP) method in which proportions are based on forecasts rather than on historical data.
It first generates an independent base forecast for all series in the hierarchy, then for each level, from the top to the bottom, the proportion of each base forecast to the aggregate of all the base forecasts at that level are calculated. For a hierarchy with $K$ levels we have:

\[
p_i = \prod_{k=0}^{K-2}\frac{\hat{y}_{t, i}^{k}}{\hat{\sigma}_{t, i}^{k+1}}, \quad i=1, ..., m_{K-1}.
\]

where $\hat{y}_{t, i}^{k}$ is the base forecast of the series that corresponds to the node which is $k$ levels above node $i$, and $\hat{\sigma}_{t, i}^{k+1}$ is the sum of the base forecasts below the series that is $k$ levels above node $i$ and directly in contact with that series.

\subsection{Middle-out Approach}
The middle-out method can be seen as a combination of the top-down and bottom-up approaches. It combines ideas from both methods by starting from a middle level where forecasts are reliable. For the series above the middle level, coherent forecasts are generated using the bottom-up approach by aggregating these forecasts upwards. For the series below the middle level, coherent forecasts are generated using a top-down approach by disaggregating the middle-level forecasts downwards.

\subsection{Optimal Reconciliation}
\cite{hyndman2011optimal} propose a novel approach that provides optimal forecasts that are better than forecasts produced by either a top-down or a bottom-up approach. Their proposal is independently forecasting all series at all levels of the hierarchy and then using a linear regression model to optimally combine and reconcile these forecasts. Their approach uses a generalized least squares estimator that requires an estimate of the covariance matrix of the errors that arise due to incoherence. In a recent paper, \cite{wickramasuriya2019optimal} show that this matrix is impossible to estimate in practice, and they propose a state-of-the-art forecast reconciliation approach, called Minimum Trace (MinT) that incorporates the information from a full covariance matrix of forecast errors in obtaining a set of coherent forecasts. MinT minimizes the mean squared error of the coherent forecasts across the entire hierarchy with the constraint of unbiasedness. The resulting revised forecasts are coherent, unbiased, and have minimum variance amongst all combination forecasts. An advantage of the optimal reconciliation approach is that allows for the correlations between the series at each level using all the available information within the hierarchy. However, it is computationally expensive compared to the other methods introduced so far because it requires to individually forecast the time series at all the levels.

\section{Neural Network Disaggregation}
\label{sec:nnd}
According to \cite{hyndman_forecasting}, standard top-down approaches have the disadvantage of information loss since they are unable to capture the individual time series characteristics. On the other hand, the bottom-up approach does not exploit the characteristics of the time series at intermediate levels. Departing from the related literature to the best of our knowledge, we propose a new approach that first generates an accurate forecast for the aggregated time series at a chosen level of the hierarchy and then disaggregates it downwards. We formulate the disaggregation problem as a non-linear regression problem, and we solve it with a deep neural network that jointly learns how to disaggregate and generate coherent forecasts across the levels of the hierarchy.\par
\begin{figure}
\centering
\begin{tikzpicture}[level distance=2cm,
level 1/.style={sibling distance=6cm},
level 2/.style={sibling distance=4cm},
level 3/.style={sibling distance=1cm}]
\tikzstyle{every node}=[circle,draw=none]

\node (Root) {$\ldots$}
    child {
    node[draw=none] {$\ldots$}
}
child {
    node {$y_{t,j}^{k,p}$} 
    child { node {$y_{t,1}^{k+1,j}$} 
    child { node[draw=none] {$\ldots$} } 
    child { node[draw=none] {$\ldots$} } 
    child { node[draw=none] {$\ldots$} } }
    child { node[draw=none] {$\ldots$} 
    child { node[draw=none] {$\ldots$} } 
    child { node[draw=none] {$\ldots$} } 
    child { node[draw=none] {$\ldots$} } }
    child { node {$y_{t,m_j^{k+1}}^{k+1,j}$} 
    child { node[draw=none] {$\ldots$} }
    child { node[draw=none] {$\ldots$} } 
    child { node[draw=none] {$\ldots$} } }
}
child {
    node[draw=none] {$\ldots$}
};

\draw (-5, -5.2) rectangle (5, -1)node[midway] {};
\end{tikzpicture}
\caption{A top-level series at level $k$ and the bottom-level series at level $k+1$.}
\label{fig:hierarchy}
\end{figure}
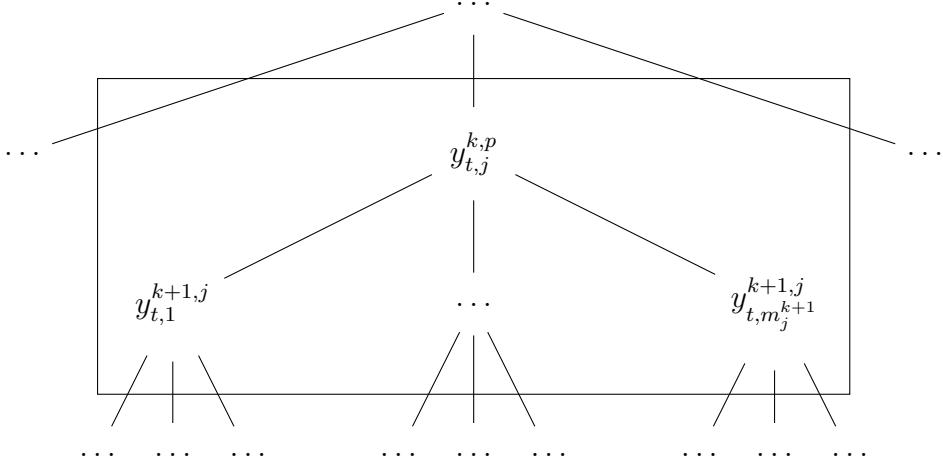

To explain the proposed algorithm, we focus on two consecutive levels with the top-level time series being at node $j$ of level $k$ and the bottom-level series at level $k+1$ (see Figure \ref{fig:hierarchy}). 

Let $m_j^{k+1}$ be the number of series at level $k+1$ connected to the parent node $j$ at level $k$, then we model the disaggregation procedure as a non-linear regression problem:

\begin{equation}
    \boldsymbol{y_{t}^{k+1, j}} = f(y_{t, j}^{k, p}, y_{t-1, j}^{k, p}, \dots, y_{t-l, j}^{k, p}, \boldsymbol{x_{t, 1}}, \dots, \boldsymbol{x_{t, m_j^{k+1}}})+\boldsymbol{\epsilon},
\end{equation}

where $\boldsymbol{y_{t}^{k+1, j}}$ is the vector of size $m_j^{k+1}$ containing the series at level $k+1$, $y_{t, j}^{k, p}$ is the aggregate time series corresponding to the node $j$ at level $k$ connected to the parent node $p$ at level $k-1$, $l$ is the number of the lagged time steps of the aggregated series, $\boldsymbol{x_{t, i}}$ is a vector of the external regressors for each series at level $k+1$, $f$ is a non-linear function learned, in our case, by a feed-forward neural network and $\boldsymbol{\epsilon}$ is the error term.

\begin{figure}
\begin{tikzpicture}[scale=1.7]
 \draw (1.5,1.75) rectangle (3.5,1)node[midway]{\small $\begin{array}{c}y_{t, j}^{k, p}\\ \mbox{Training Set}\end{array} $};
 \draw (5.5,1.75) rectangle (7.5,1)node[midway]{\small $\begin{array}{c}\boldsymbol{y_{t}^{k+1, j}, x_{t, i}}\\ \mbox{Training Set}\end{array} $};
 \draw (1.5,0) rectangle (3.5,-0.75)node[midway]{\small $\begin{array}{c}\mbox{Model Selection} \end{array} $};
 \draw (5.5,-0) rectangle (7.5,-0.75)node[midway]{\small NND Training};
 \draw (2.5,-2.5) ellipse (1 and 0.35)node{\small $F^*$ Model};
 \draw (1.5,-5.2) rectangle (3.5,-4.5)node[midway]{\small $\hat{y}_{t, j}^{k, p}$ Test Set};
 \draw (6.5,-4.85) ellipse (1 and 0.4)node{\small NND Model};
  \draw (5.5,-6) rectangle (7.5,-6.7)node[midway]{\small $\boldsymbol{\hat{y}_{t}^{k+1, j}}$ Test Set};
\draw[->](2.5,1)--(2.5,0);
\draw[->](2.5,-0.75)--(2.5,-2.15);
\draw[->](2.5,-2.85)--(2.5,-4.5);

\draw[->](6.5,1)--(6.5,0);
\draw[->](6.5,-0.75)--(6.5,-4.45);
\draw[->](6.5,-5.25)--(6.5,-6);

\draw[-](3.5,1.375)--(4.5,1.375);
\draw[->](4.5,1.375)--(4.5,-0.375)--(5.5,-0.375);

\draw[->](3.5,-4.85)--(5.5,-4.85);

\draw (0.5,0.5) rectangle (8.5,-3.25)node[midway] {\bf Step 1};
\draw (0.5,-3.75) rectangle (8.5,-7.25)node[midway] {\bf Step 2};\end{tikzpicture}
\caption{Decomposition of the aggregated forecast through a neural network: Neural Network Disaggregation (NND).}
 \label{fig:algorithm}
\end{figure}
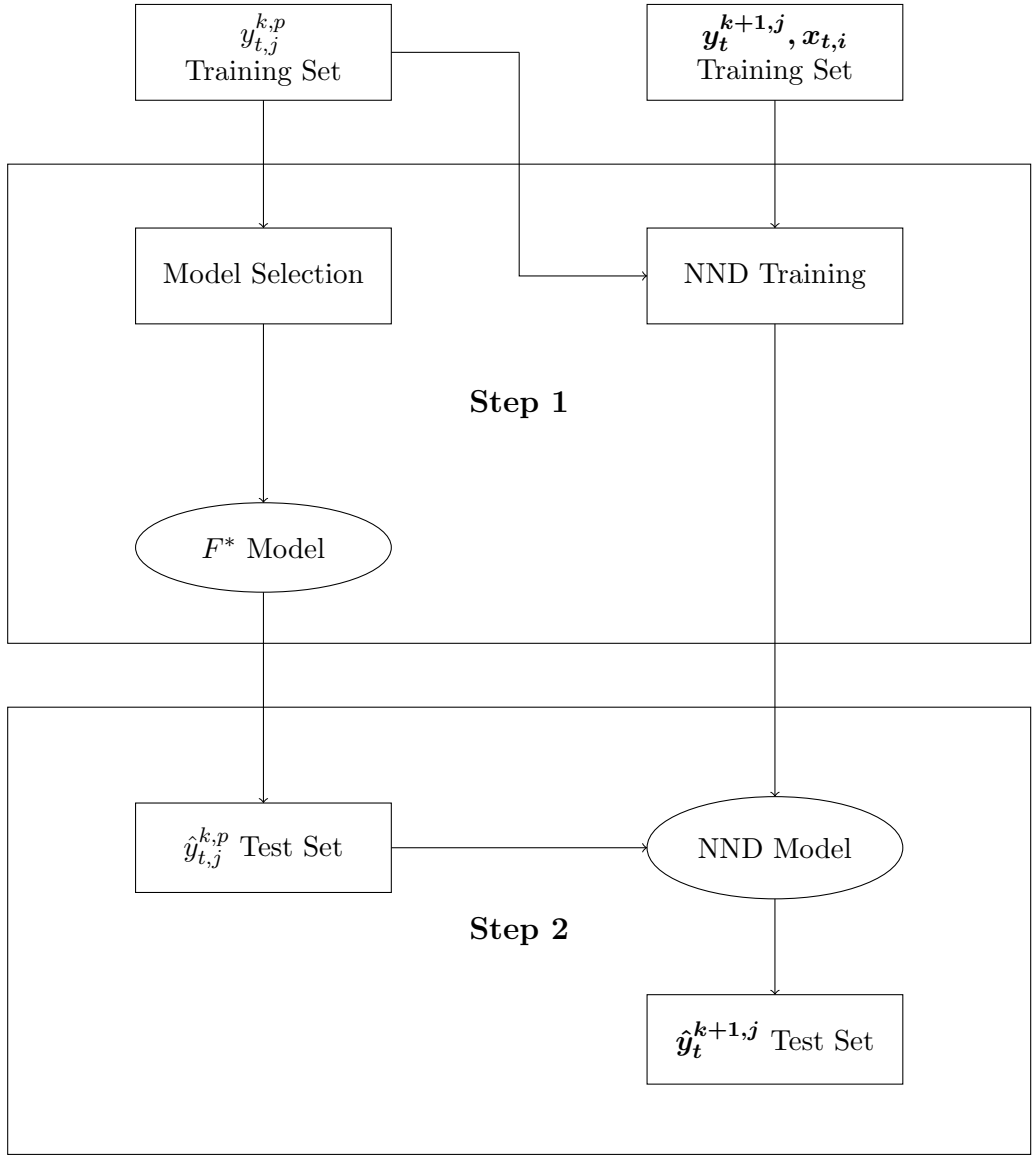

Given any aggregate time series $y_{t, j}^{k, p}$ and the vector of series $\boldsymbol{y_{t}^{k+1, j}}$, the algorithm is made up of two steps. In the first one, the best forecasting model for the aggregated time series is chosen, and the neural network is trained with the real values of the training set of the two levels time series. In the second step, forecasts for the aggregated time series are fed to the neural network to obtain forecasts for all the lower-level time series. The flow chart of the proposed algorithm is shown in Figure \ref{fig:algorithm}. More in detail, the two steps are the following:
\begin{enumerate}
    \item[Step 1] In the training phase, the best forecasting model $F^*$ for the time series $y_{t, j}^{k, p}$ is chosen based on the training set. At the same time, the neural network is trained taking as input the training set of $y_{t, j}^{k, p}$ with lagged time steps and the explanatory variables $\boldsymbol{x_{t, i}}$ relative to the training set of $\boldsymbol{y_{t}^{k+1, j}}$. The output are the true values of the disaggregated time series $\boldsymbol{y_{t}^{k+1, j}}$. In order to simplify the notation, from now on we refer to the produced model as NND (Neural Network Disaggregation).
    \item[Step 2] In the disaggregation or test phase, forecasts $\hat{y}_{t, j}^{k, p}$ relative to the time period of the test set are generated by the model $F^*$. Finally, these forecasts are fed to the trained NND to produce the disaggregated forecasts $\boldsymbol{\hat{y}_{t}^{k+1, j}}$ for the test set.
\end{enumerate}

In general, the learned function $f$ generates base forecasts that are not coherent since they do not sum up correctly according to the hierarchical structure. In order to ensure that forecasts are reconciled across the hierarchy, we want $f$ to output a set of forecasts that are as close as possible to the base forecasts, but also meet the requirement that forecasts at upper levels in the hierarchy are the sum of the associated lower-level forecasts. From an optimization perspective, we want to introduce an equality constraint to the regression problem in such a way that we can still use backpropagation to train the network. More in detail, we are looking for the network weights such that the fitting error is minimized and, besides, we want the following constraint to hold:

\begin{equation}
    y_{t, j}^{k, p} = \boldsymbol{1^\mathsf{T} y_{t}^{k+1, j}} = \boldsymbol{1^\mathsf{T} \hat{y}_{t}^{k+1, j}} = \hat{y}_{t, j}^{k, p},
\end{equation}

where $\boldsymbol{1}$ is the vector of all ones of size $m_j^{k+1}$. 

We impose the coherence by adding a term to the fitting error that penalizes differences between the sum of the lower-level observations and the sum of the lower-level forecasts:

\begin{equation}\label{loss}
\begin{split}
L(\boldsymbol{y_{t}^{k+1, j}}, \boldsymbol{\hat{y}_{t}^{k+1, j}}) = \frac{1}{T} \bigg[ (1-\alpha) \sum_{t = 1}^{T} ||\boldsymbol{y_{t}^{k+1, j}} - \boldsymbol{\hat{y}_{t}^{k+1, j}}||^2 + \\ + \alpha \sum_{t = 1}^{T} (\boldsymbol{1^\mathsf{T} y_{t}^{k+1, j}} - \boldsymbol{1^\mathsf{T} \hat{y}_{t}^{k+1, j}})^2\bigg],
\end{split}
\end{equation}
where $\alpha \in (0, 1)$ is a parameter that controls the relative contribution of each term in the loss function. Note that the two terms are on the same scale and the parameter $\alpha$ measures the compromise between minimizing the fitting error and satisfying the coherence. A too small value of $\alpha$ will result in the corresponding constraint being ignored, producing, in general, not coherent forecasts whereas a too large value will cause the fitting error being ignored, producing coherent but possibly inaccurate base forecasts. The idea is to balance the contribution of both terms by setting $\alpha = 0.5$, that corresponds to giving the two terms the same importance. 
In principle, the parameter $\alpha$ may be tuned on each instance. However, we did not investigate the tuning of $\alpha$ and kept it fixed to $0.5$, since this setting allowed us to reach a satisfying reconciliation error on all the experiments.

Top-down approaches distribute the top-level forecasts down the hierarchy using historical or forecasted proportions of the data. In our case, explicit proportions are never calculated since the algorithm automatically learns how to disaggregate forecasts from any level of the hierarchy to the series below without loss of information. Furthermore, our method is flexible enough to be employed in the forecasting process of the whole hierarchy in two different ways:
\begin{enumerate}
    \item {\it Standard top-down:} a forecasting model $F^*$ is developed for the aggregate at level $0$, and a single disaggregation model NDD is trained with the series at level $0$ and $K-1$. Therefore, forecasts for the bottom-level series are produced by looking only at the aggregated series at level $0$. Then, the bottom-level forecasts are aggregated to generate coherent forecasts for the rest of the series of the hierarchy.
    \item {\it Iterative top-down:} the forecasting model $F^*$ for an aggregate at level $k$ is the disaggregation model NDD trained with the series at level $k-1$ and $k$, for each $k=1, \dots, K-1$. At level $0$, instead, $F^*$ is the best model selected among a set of standard forecasting methods. Forecasts for all the levels are then obtained by feeding forecasts to the disaggregation models at each level. 
\end{enumerate}
The difference between the two approaches is that in the standard top-down, bottom-level forecasts are generated with only one disaggregation model, whereas in the iterative version, a larger number of disaggregation models is trained, one for each series to be disaggregated. To be more precise, to disaggregate the $m_k$ series at level $k=0,\dots,K-2$, exactly $m_k$ disaggregation models are trained in parallel. In this way, on the one hand, we increase the variance of the approach (and the computational time), but on the other hand, we reduce the bias since we increase flexibility and keep into account more the variability at the different levels.

We also notice that this algorithm can be easily plugged into a middle-out strategy: a forecasting model is developed for each aggregate at a convenient level, and the disaggregation models are trained and tested to distribute these forecasts to the series below. For the series above the middle level, coherent forecasts are generated using the bottom-up approach.

\begin{figure}
\centering
\begin{tikzpicture}[scale=1.75]

\draw (0,0) rectangle (2,1)node[midway]{\small $\begin{array}{c} \mbox{Multi-Layer} \\ \mbox{Perceptron} \\ \mbox{(MLP)}\end{array} $};
\draw (4,0) rectangle (6,1)node[midway]{\small $\begin{array}{c} \mbox{Convolutional} \\ \mbox{Neural Network} \\ \mbox{(CNN)}\end{array} $};
\draw (1,3) ellipse (1 and 0.5)node{\small $\begin{array}{c} \mbox{Explanatory}\\ \mbox{Variables}\end{array} $};
\draw (5,3) ellipse (1 and 0.5)node{\small $\begin{array}{c} \mbox{Time Series}\\ \mbox{Data}\end{array} $};
\draw (4,-2.7) rectangle (2,-1.7)node[midway]{\small Concatenate};
\draw (4,-3.7) rectangle (2,-4.7)node[midway]{\small $\begin{array}{c} \mbox{Fully-Connected} \\ \mbox{(Linear Activation)}\end{array} $};

\draw[->](5,2.5)--(5,1);
\draw[->](1,2.5)--(1,1);
\draw[-](5,-1)--(1,-1);
\draw[-](5,-0)--(5,-1);
\draw[-](1,0)--(1,-1);
\draw[->](3,-1)--(3,-1.7);
\draw[->](3,-2.7)--(3,-3.7);

\draw (-0.5, 4) rectangle (2.5,-0.70)node[above left] { \bf Branch 1};
\draw (3.5, 4) rectangle (6.5,-0.70)node[above left] { \bf Branch 2};

\end{tikzpicture}
\caption{Our model has one branch that accepts the numerical data (left) and another branch that accepts time series data (right).}
\label{fig:model}
\end{figure}
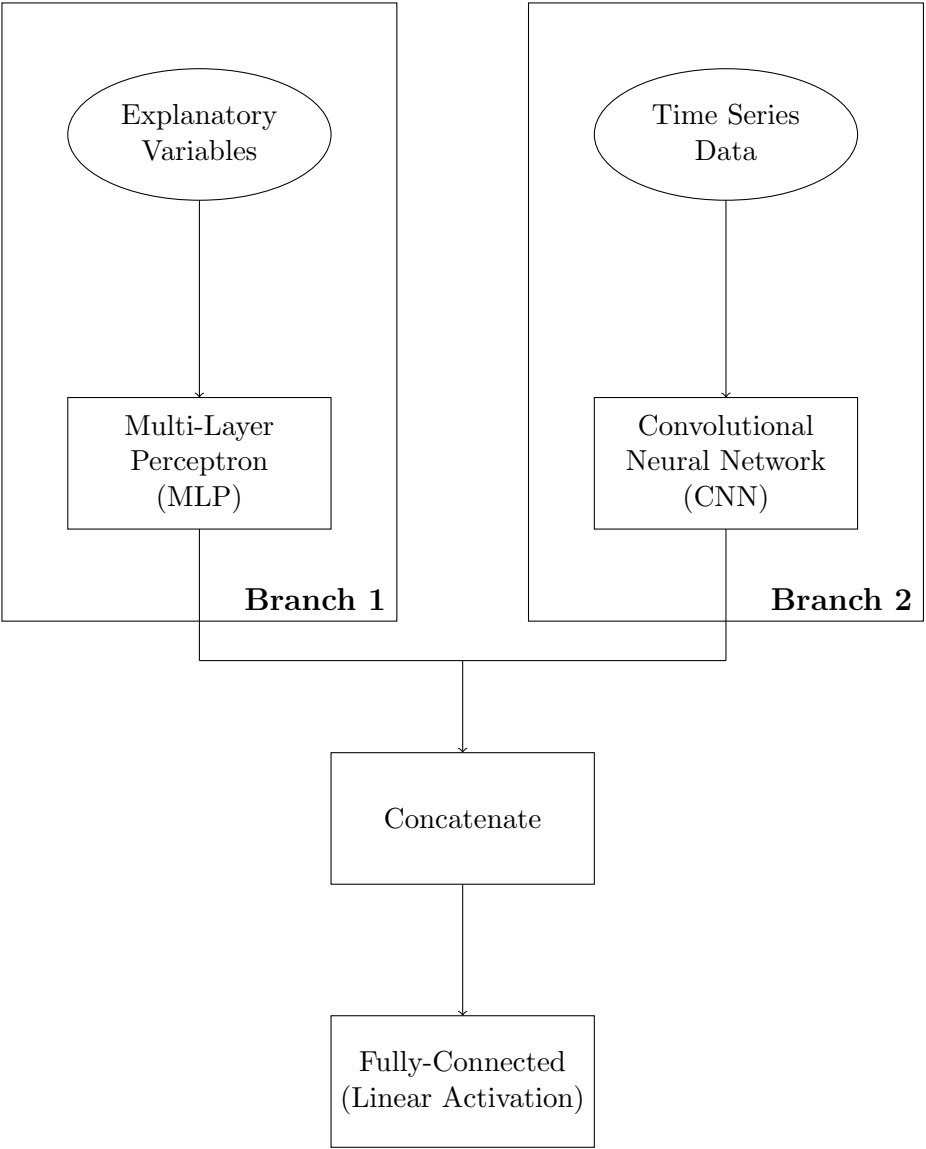

Regarding the choice of the neural network architecture, our objective is to include in the model the relationship between explanatory variables derived from the lower-level series, and the features of the aggregate series that describe the structure of the hierarchy. In order to better capture the structure of the hierarchy, we use a Convolutional Neural Network (CNN). CNNs are well known for creating useful representations of time series automatically, being highly noise-resistant models, and being able to extract very informative, deep features, which are independent of time \citep{kanarachos2017detecting, ferreira2018designing}. Our model is a deep neural network capable of accepting and combining multiple types of input, including cross-sectional and time series data, in a single end-to-end model. Our architecture is made up of two branches: the first branch is a simple Multi-Layer Perceptron (MLP) designed to handle the explanatory variables $\boldsymbol{x_{t, i}}$ such as, promotions, day of the week, or in general, special events affecting the time series of interest; the second branch is a one-dimensional CNN that extracts feature maps over fixed segments of length $w$ from the aggregate series $y_{t, j}^{k, p}$.  Features extracted from the two subnetworks are then concatenated together to form the final input of the multi-output regression model (see Figure \ref{fig:model}). The output layer of the model is a standard regression layer with linear activation function where the number of units is equal to the number of the series to forecast.

\section{Experimental Setup}
\label{sec:setup}

In this section, we resume first the forecasting models we use to generate the base forecasts for the hierarchical approaches, then we describe our strategy to select the best forecasting model and the implementation details. 

\subsection{Forecasting Models}
In order to describe the methods, let $(y_1, \dots, y_T)$ be an univariate time series of length $T$ and $(y_{T+1}, \dots, y_{T+h})$ the forecasting period, where $h$ is the forecast horizon. We consider the following models:

\begin{enumerate}

    \item {\it Naive}
    \item {\it Autoregressive Integrated Moving Average (ARIMA)}
    \item {\it Exponential Smoothing (ETS)}
    \item  {\it Non-linear autoregression model (NAR) }
    \item {\it Dynamic regression models}: univariate time series models, such as linear and non-linear autoregressive models, allow for the inclusion of information from past observations of a series, but not for the inclusion of other information that may also affect the time series of interest. Dynamic regression models allow keeping into account the time-lagged relationship between the output and the lagged observations of both the time series itself and of the external regressors. More in detail, we consider two types of dynamic regression models:
    \begin{enumerate}
        \item {\it ARIMA model with exogenous variables (ARIMAX)}
        \item {\it NAR model with exogenous variables (NARX)}
    \end{enumerate}

\end{enumerate}
In the literature, it has been pointed out that the performance of forecasting models could be improved by suitably combining forecasts from standard approaches \citep{combinations}. An easy way to improve forecast accuracy is to use several different models on the same time series and to average the resulting forecasts. We consider two ways of combining forecasts:

\begin{enumerate}
    \item {\it  Simple Average:} the most natural approach to combine forecasts is to use the mean. The composite forecast in case of simple average is given by $\hat{y}_t = \frac{1}{m}\sum_{i=1}^{m} \hat{y}_{i,t}$ for $t=T+1, ..., T+h$ where $h$ is the forecast horizon, $m$ is the number of combined models and $\hat{y}_{i, t}$ is the forecast at time $t$ generated by model $i$.
\item {\it Constrained Least Squares Regression:} the composed forecast is not a function of $m$ only as in the simple average but is a linear function of the individual forecasts whereby the parameters are determined by solving an optimization problem. The approach proposed by \cite{combinations} minimizes the sum of squared errors under some additional constraints. Specifically, the estimated coefficients $\beta_i$ are constrained to be non-negative and to sum up to one. The weights obtained are easily interpretable as percentages devoted to each of the individual forecasts. Given the optimal weights, the composed forecast is obtained as $\hat{y}_t=\sum_{i=1}^{m}\beta_i \hat{y}_{i, t}$ for $t=T+1, ..., T+h$. From the mathematical point of view the following optimization problem needs to be solved:

\begin{align}\nonumber
               \min &\ \sum_{t=T+1}^{T+h}(y_t - \sum_{i=1}^{m}\beta_i \hat{y}_{i, t})^2 \ & \\
            \label{CLSv1}   
\textrm{s.t.}\       &\ \beta_i \geq 0 &   i = 1, \ldots, m \\
           \nonumber
            &\ \sum_{i=1}^{m}\beta_i = 1        
\end{align}

Different from the simple average which does not need any training as the weights are a function of $m$ only, with this method we need to allocate a reserved portion of forecasts to train the meta-model. 
\end{enumerate}

In particular, we consider two following composite models:
\begin{enumerate}
    \item Combination of ARIMAX, NARX, and ETS forecasts obtained through the simple mean.
    \item Combination of ARIMAX, NARX, and ETS forecasts obtained by solving the constrained least squares problem.
\end{enumerate}

We choose to combine these two dynamic regression models with exponential smoothing in order to take directly into account the effect of the explanatory variables and the presence of linear and non-linear patterns in the series.

\subsection{Model Selection}
Following an approach widely employed in the machine learning literature, we separate the available data into two sets, training (in-sample) and test (out-of-sample) data. The training data $(y_1, \dots, y_N)$, a time series of length $N$, is used to estimate the parameters of a forecasting model and the test data $(y_{N+1}, \dots, y_{T})$, that comes chronologically after the training set, is used to evaluate its accuracy.

\begin{figure}
  \centering
  \includegraphics[width=\textwidth]{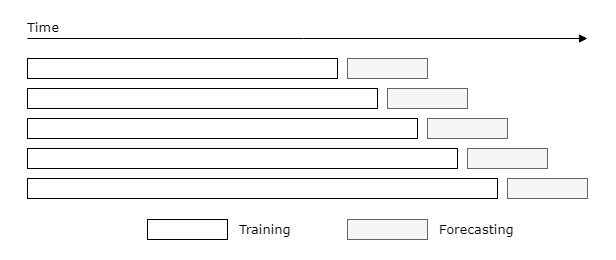}
 \caption{Expanding window procedure.}
 \label{fig:expanding}
\end{figure}

To achieve a reliable measure of model performance, we implement on the training set a procedure that applies a cross-validation logic suitable for time series data. In the expanding window procedure described by \cite{hyndman_forecasting}, the model is trained on a window that expands over the entire history of the time series, and it is repeatedly tested against a forecasting window without dropping older data points. This method produces many different train/test splits, and the error on each split is averaged in order to compute a robust estimate of the model error (see Figure \ref{fig:expanding}). The implementation of the expanding window procedure requires four parameters:

\begin{itemize}
  \item[-] \textit{Starting window:} the number of data points included in the first training iteration.
  \item[-] \textit{Ending window:} the number of data points included in the last training iteration. 
  \item[-] \textit{Forecasting window:} number of data points included for forecasting.
  \item[-] \textit{Expanding steps:} the number of data points added to the training time series from one iteration to another.
\end{itemize}

For each series, the best performing model after the cross-validation phase is retrained using the in-sample data, and forecasts are obtained recursively over the out-of-sample period. The above procedure requires a forecast error measure. We consider the Mean Absolute Scaled Error (MASE) proposed by \cite{hyndman2006another}:
\[
MASE = \frac{\frac{1}{h} \sum_{i=T+1}^{T+h}|y_{i} - \hat{y}_{i}|}{\frac{1}{T-m}\sum_{t=m+1}^{T} |y_t - y_{t-m}|},
\]
where the numerator is out-of-sample Mean Absolute Error (MAE) of the method evaluated across the forecast horizon $h$, and the denominator is the in-sample one-step ahead Naive forecast with seasonal period $m$.

We also consider the Symmetric Mean Absolute Percentage Error (SMAPE) defined as follows:
\[
SMAPE = \frac{2}{h} \sum_{i=T+1}^{T+h} \frac{ |y_{i} - \hat{y}_{i}| }{|y_{i}| + |\hat{y}_{i}|}.
\]
The SMAPE is easy to interpret, and has an upper bound of 2 when either actual or predicted values are zero or when actual and predicted are opposite signs. However, the significant disadvantage of SMAPE is that it produces infinite or undefined values where the actual values are zero or close to zero.
The MASE and SMAPE can be used to compare forecast methods on a single series and, because they are scale-free, to compare forecast accuracy across series. For this reason, we average the MASE and SMAPE values of several series to obtain a measurement of forecast accuracy for the group of series.

\subsection{Implementation}
Time series models described above are implemented by using the \texttt{forecast} package in R \citep{forecast_package}. Hierarchical time series forecasting is performed with the help of \texttt{hts} package in R \citep{hts}. For the optimal reconciliation approach, we use the MinT algorithm that estimates the covariance matrix of the base forecast errors using shrinkage \citep{wickramasuriya2019optimal}. The proposed NND is implemented in Python with \texttt{TensorFlow}, a large-scale machine learning library \citep{tensorflow}. The CNN subnetwork has 6 convolutional layers with ReLU activation, whereas the MLP subnetwork has 3 fully connected layers with ReLU activation.
The hyperparameters optimization of the CNN subnetwork regards the number of filters (F) and the kernel size (K) of the convolutional layers, whereas the optimization of the MLP subnetwork regards the number of units in the hidden layers (H). Grid search is used to perform the hyperparameters optimization search in the space of the neural network where $F=\{16, 32, 64\}$, $K=\{4, 8, 16\}$ and $H=\{64, 128, 256\}$. We evaluate the hyperparameters configuration on a held-out validation set, and we choose the architecture achieving the best performance on it (see \ref{sec:app1} for the optimal hyperparameters of the trained models). The NND model takes as input mini-batches of 32 examples and the loss function in equation \eqref{loss} is minimized by using the Adam optimizer \citep{adam} with the initial learning rate set to 0.001. The network is trained for 500 epochs, and early stopping is used to stop the training as soon as the error on the validation set starts to grow \citep{early_stopping}. The training time of a single disaggregation model requires order of minutes on the Intel Core i7-8565U CPU depending on the network dimension and on the granularity of the dataset.

\section{Numerical Experiments}
\label{sec:exp}

In this section, we aim to evaluate the effectiveness of our approach by comparing it with the hierarchical methods described in Section \ref{sec:hier}. These methods are used for benchmarking the proposed method, as they
have been successfully applied in numerous applications and are considered the state-of-the-art in the area of hierarchical forecasting \citep{hollyman2021}. In order to be as fair as possible in the comparison, we perform model selection among the set of forecasting methods described in Section \ref{sec:setup} whenever a base forecast is required. This means that different methods may be used for each time series of the hierarchy we are trying to forecast (bottom-level series for the bottom-up approach, the top-level series for all the top-down, all the time series for the optimal reconciliation approach). Note that improving the accuracy of the base forecasts using the cross-validation procedure described in Section \ref{sec:setup} is also beneficial for the competitors. As for the metrics used for comparison, we use both the MASE and the SMAPE where possible (i.e., where no zeros are present). 

\subsection{Datasets}

\begin{enumerate}
\item \textbf{Italian Dataset:} we consider sales data gathered from an Italian grocery store\footnote{\url{https://data.mendeley.com/datasets/s8dgbs3rng/1}} \citep{mancuso2021}. The dataset consists of 118 daily time series representing the demand for pasta from 01/01/2014 to 31/12/2018. Besides univariate time series data, the quantity sold is integrated by information on the presence or the absence of a promotion (no detail on the type of promotion on the final price is available). These time series can be naturally arranged to follow a hierarchical structure. Here, the idea is to build a 3-level structure: at the top of the hierarchy, there is the total or the store-level series obtained by aggregating the brand-level series. At the second level, there are the brand-level series (like for instance Barilla) obtained by aggregating the individual demand at the item level. Finally, the third level contains the most disaggregated time series representing the item-level demand (for example the demand for spaghetti Barilla). The completely aggregated series at level 0 is disaggregated into 4 component series at level 1 (B1 to B4). Each of these series is further subdivided into 42, 45, 10, and 21 series at level 2, the completely disaggregated bottom level representing the different varieties of pasta for each brand (see Table \ref{tab:hierarchy}).
\item \textbf{Electricity Dataset}: we use a public electricity demand dataset that contains power measurements and meteorological forecasts relative to a set of 24 power meters installed in low-voltage cabinets of the distribution network of the city of Rolle in Switzerland \citep{nespoli2020hierarchical}. The dataset contains measurements from 13/01/2018 to 19/01/2019 at the granularity of 10 minutes and includes mean active and reactive power, voltage magnitude, maximum total harmonic distortion for each phase, voltage frequency, and the average power over the three phases. We assume that the grid losses are not significant, so the power at the grid connection is the algebraic sum of the connected nodes. Based on the historical measurements, the operator can determine coherent forecasts for all the grid by generating forecasts for the nodal injections individually. We build a 2-level hierarchy in which we aggregate the 24 series (M1 to M24) of the distribution system at the meter level to generate the total series at the grid level (see Table \ref{tab:hierarchy_electricity}).
\item \textbf{Walmart Dataset}: we consider a public dataset made available by Walmart that was adopted in the latest M competition, M5\footnote{\url{https://mofc.unic.ac.cy/m5-competition/}}. This dataset contains historical sales data from 29/01/2011 to 19/06/2016 of various products sold in the USA, organized in the form of grouped time series. More specifically, it uses unit sales data collected at the product-store level that are grouped according to product departments, product categories, stores, and three geographical areas: the States of California (CA), Texas (TX), and Wisconsin (WI). Besides the time series data, it includes explanatory variables such as promotions (SNAP events), days of the week, and special events (e.g., Super Bowl, Valentine’s Day, Thanksgiving Day) that typically affect unit sales and could improve forecasting accuracy. Starting from this dataset, we extract a 4-level hierarchy: the completely aggregated series at level 0 is divided into 3 component series at level 1 representing the state-level time series (CA, TX, WI). The state-level time series are respectively subdivided in 4 (CA1, CA2, CA3, CA4), 3 (TX1, TX2, TX3), and 3 (WI1, WI2, WI3) time series at level 2, the store level. Finally, each store-level time series is further subdivided into 3 time series at the category level, the most disaggregated one, containing the categories Foods, Hobbies, and Household (see Table \ref{tab:hierarchy_m5}).
\end{enumerate}

\begin{table}
\centering
\begin{tabular}{ |c|c c| } 
 \hline
 \textbf{Level} & \textbf{Number of series} & \textbf{Total series per level} \\ 
 \hline
 Store & 1 & 1 \\ 
 \hline
 Brand & 4 & 4 \\ 
 \hline
 Item & 42 - 45 - 10 - 21 & 118 \\
 \hline

\end{tabular}
\caption{Hierarchy for the Italian sales data.}
\label{tab:hierarchy}
\end{table}

\begin{table}
\centering
\begin{tabular}{ |c|c c| } 
 \hline
 \textbf{Level} & \textbf{Number of series} & \textbf{Total series per level} \\ 
 \hline
 Grid & 1 & 1 \\ 
 \hline
 Meter & 24 & 24 \\ 
 \hline

\end{tabular}
\caption{Hierarchy for the electricity demand data.}
\label{tab:hierarchy_electricity}
\end{table}

\begin{table}
\centering
\begin{tabular}{ |c|c c| } 
 \hline
 \textbf{Level} & \textbf{Number of series} & \textbf{Total series per level} \\ 
 \hline
 Total & 1 & 1 \\ 
 \hline
 State & 3 & 3 \\ 
 \hline
 Store & 4 - 3 - 3 & 10 \\
 \hline
 Category & 3 - 3 - 3 - 3 - 3 - 3 - 3 - 3 - 3 - 3 & 30 \\
 \hline

\end{tabular}
\caption{Hierarchy for the Walmart data.}
\label{tab:hierarchy_m5}
\end{table}

To summarize, we have the first dataset with a three-level hierarchy, the second one with a two-level hierarchy, and the third one with a four-level hierarchy. As for the experimental setup, we have to make some choices for each dataset:
\begin{enumerate}
\item \textbf{Italian Dataset:} for each series, as explanatory variables, we add a binary variable representing the presence of promotion if the disaggregation is computed at the item level or a variable representing the relative number of items in promotion for each brand if the disaggregation is computed at the brand level. In both cases, dummy variables representing the day of the week and the month are also added to the model. As for the number of lagged observations of the aggregate demand, we consider time windows of length $w=30$ days with a hop size of 1 day. 
We consider 4 years from 01/01/2014 to 31/12/2017 for the in-sample period and the last year of data from 01/01/2018 to 31/12/2018 for the out-of-sample period. The experimental setup for the cross-validation procedure is as follows. The starting window consists of the first three years of data from 01/01/2014 to 31/12/2016. The training window expands over the last year of the training data including daily observations from 01/01/2017 to 31/12/2018. The forecasting window is set to $h=7$, corresponding to a forecasting horizon of one week ahead. At each iteration, the training window expands by one week to simulate a production environment in which the model is re-estimated as soon as new data are available and to better mimic the practical scenario in which retailing decisions occur every week. To evaluate the forecasting accuracy at each level, for this hierarchy we use the average MASE, as recommended by \cite{hyndman2006another}, since most of the item-level series are intermittent.
\item \textbf{Electricity Dataset:} for each series, we use the average power over the three phases as the target variable and the temperature, horizontal irradiance, normal irradiance, relative humidity, pressure, wind speed, and wind direction as explanatory variables. Dummy variables representing the day of the week and the hour of the day are also added to the model. 
As for the number of lagged observations of the aggregate power, we consider time windows of length $w=144$ observations with the hop size of 10 minutes. We consider 9 months from 13/01/2018 to 13/09/2018 for the training set and the last 3 months from 14/09/2018 to 13/01/2019 for the test set. The configuration of the cross-validation procedure is as follows. The starting window consists of the first six months of data from 13/01/2018 to 13/06/2018. At each iteration, the training window expands by 24 hours over the last 3 months of the training data including observations from 14/06/2018 to 13/09/2018. The forecasting window is set to $h=144$, corresponding to a forecasting horizon of 24 hours ahead. We evaluate the forecasting accuracy at each level by using the average MASE and the average SMAPE over all the series of that level since there are no zero values in these time series.
\item \textbf{Walmart Dataset:} for each series, we use dummy variables representing the day of the week and the month. We also include as explanatory variables snap events and special events affecting sales: National holidays, Religious holidays, Sporting events, Valentine's Day, Father's Day, and Mother's Day. We consider 4 years from 29/01/2011 to 29/06/2015 for the in-sample period and the last year of data from 30/06/2015 to 16/06/2016 for the out-of-sample period. The experimental setup for the cross-validation procedure is as follows. The starting window consists of the first three years of data from 29/01/2011 to 29/01/2014. The training window iteratively expands over the last year of the training data including observations from 30/01/2014 to 29/06/2015. The forecasting horizon is set to $h = 7$, and the number of lagged observations for the aggregate sales is set to $w=30$ days with the hop size of 1 day. We evaluate the quality of the forecasts by looking at the MASE since the SMAPE is undefined due to the presence of zeros values when Walmart is closed.
\end{enumerate}

\subsection{Results}
We compare the forecasting performance of our method for each series, in both its versions, standard top-down (NND1) and iterative top-down (NND2) with the bottom-up (BU), average historical proportions (AHP), proportions of historical averages (PHA), forecasted proportions (FP) and the optimal reconciliation approach through trace minimization (OPT). We stress that for all the top-down approaches, the performance at the most aggregated level is equivalent, and the differences only emerge at the lower levels of the hierarchy, where we are interested in the comparison.

For all the datasets, we report the metrics on all the considered time series, comparing also the average error at each level. To formally test whether the forecasts produced by the considered hierarchical methods are different, we use the non-parametric Friedman and post-hoc Nemenyi tests as in \cite{koning2005m3}, \cite{demvsar2006statistical}, and \cite{di2020cross}. As stated by \cite{kourentzes2019cross}, the Friedman test first establishes whether at least one of the forecasts is statistically different from the others. If this is the case, the Nemenyi test identifies groups of forecasts for which there is no evidence of significant differences. The advantage of this approach is that it does not impose any assumption on the distribution of the data and does not require multiple pairwise tests between forecasts, which would distort the outcome of the tests. The hierarchical forecasting methods are then sorted according to the mean rank with respect to the considered metric.\par

In Table \ref{tab:results_pasta} we provide results for all 123 time series of the Italian sales dataset. For the NND1, we directly forecast the demand at the item level using the aggregate demand at the store level, and then we aggregate the item-level forecasts to obtain the brand-level forecasts. For the NND2, we train a disaggregation model that generates the brand-level forecasts starting from the store-level series, and then one NND for each brand-level series to generate forecasts for each item demand of the brand they belong to. Overall, for the entire hierarchy we train one NND at the top level, and 4 NND in parallel at the brand level.

\begin{tabularx}{\textwidth}{lccccccc}
\caption{MASE for all 123 series of the Italian sales dataset. In bold the best performing approach.} \\
\toprule
{} &     BU &  AHP &  PHA &   FP &   NND1 &   NND2 &   OPT \\
\midrule
TOTAL  &  0.999 &  0.562 &  0.562 &  0.562 &  0.562 &  0.562 &  \bf{0.560} \\
\midrule
B1     &  1.321 &  1.213 &  1.225 &  1.249 &  0.752 &  \bf{0.702} &  0.856 \\
B2     &  0.729 &  1.009 &  1.027 &  0.740 &  0.788 &  \bf{0.764} &  0.776 \\
B3     &  1.265 &  1.749 &  1.910 &  1.225 &  0.772 &  \bf{0.677} &  0.877 \\
B4     &  1.443 &  1.610 &  1.660 &  1.325 & \bf{0.723} &  0.737 &  0.926 \\
\midrule
Average Brand & 1.189	& 1.395	& 1.455 &	1.135 & 0.759 & \bf{0.720} & 0.859 \\
\midrule
B1-I1   &  1.295 &  1.100 &  1.099 &  1.360 &  0.669 &  \bf{0.605} &  1.292 \\
B1-I2   &  1.011 &  0.930 &  0.931 &  1.024 &  \bf{0.723} &  0.747 &  1.007 \\
B1-I3   &  0.870 &  0.832 &  0.820 &  0.793 &  0.795 &  \bf{0.677} &  0.769 \\
B1-I4   &  0.753 &  0.874 &  0.819 &  0.753 &  \bf{0.644} &  0.666 &  0.731 \\
B1-I5   &  1.483 &  1.414 &  1.419 &  1.589 &  \bf{0.604} &  0.639 &  0.915 \\
B1-I6   &  0.886 &  0.758 &  0.749 &  0.793 &  \bf{0.691} &  0.760 &  0.786 \\
B1-I7   &  0.838 &  0.889 &  0.849 &  0.850 &  \bf{0.659} &  0.724 &  0.832 \\
B1-I8   &  0.835 &  0.786 &  0.785 &  0.760 &  \bf{0.625} &  0.653 &  0.731 \\
B1-I9   &  1.020 &  1.032 &  1.059 &  1.044 &  0.662 &  \bf{0.626} &  0.922 \\
B1-I10  &  1.113 &  1.100 &  1.128 &  1.078 &  \bf{0.617} &  0.672 &  0.868 \\
B1-I11  &  0.944 &  0.838 &  0.836 &  0.965 &  \bf{0.691} &  0.743 &  0.951 \\
B1-I12  &  1.247 &  1.154 &  1.191 &  1.084 &  \bf{0.694} &  0.730 &  0.843 \\
B1-I13  &  0.997 &  0.810 &  0.812 &  0.921 &  \bf{0.635} &  0.638 &  0.886 \\
B1-I14  &  0.945 &  0.876 &  0.894 &  0.946 &  0.687 &  \bf{0.651} &  0.922 \\
B1-I15  &  1.090 &  1.124 &  1.125 &  1.107 &  \bf{0.632} &  0.665 &  0.801 \\
B1-I16  &  0.746 &  0.778 &  0.757 &  0.754 &  0.733 &  \bf{0.678} &  0.736 \\
B1-I17  &  0.876 &  0.803 &  0.804 &  0.901 &  0.968 &  1.027 &  \bf{0.884} \\
B1-I18  &  1.642 &  1.215 &  1.228 &  1.176 &  0.692 &  0.697 &  \bf{0.644} \\
B1-I19  &  0.846 &  0.788 &  0.757 &  0.786 &  0.760 &  0.772 &  \bf{0.738} \\
B1-I20  &  0.939 &  0.861 &  0.864 &  0.864 &  \bf{0.637} &  0.701 &  0.820 \\
B1-I21  &  0.872 &  0.777 &  0.767 &  0.778 &  \bf{0.676} &  0.705 &  0.755 \\
B1-I22  &  0.813 &  0.765 &  0.763 &  0.812 &  \bf{0.582} &  0.679 &  0.802 \\
B1-I23  &  1.054 &  0.901 &  0.907 &  1.084 &  \bf{0.502} &  0.629 &  1.057 \\
B1-I24  &  1.082 &  1.011 &  1.074 &  1.127 &  \bf{0.603} &  0.617 &  1.079 \\
B1-I25  &  0.823 &  0.905 &  0.867 &  0.834 &  \bf{0.627} &  0.681 &  0.818 \\
B1-I26  &  0.784 &  0.841 &  0.829 &  0.798 &  \bf{0.765} &  0.830 &  0.783 \\
B1-I27  &  0.747 &  0.747 &  0.725 &  0.753 &  \bf{0.678} &  0.702 &  0.743 \\
B1-I28  &  0.983 &  1.029 &  1.022 &  1.021 &  \bf{0.586} &  0.726 &  0.986 \\
B1-I29  &  1.082 &  0.874 &  0.868 &  0.889 &  \bf{0.600} &  0.692 &  0.882 \\
B1-I30  &  0.972 &  0.826 &  0.833 &  0.866 &  \bf{0.515} &  0.638 &  0.858 \\
B1-I31  &  0.955 &  0.890 &  0.896 &  0.972 &  0.690 &  \bf{0.688} &  0.938 \\
B1-I32  &  1.294 &  1.076 &  1.155 &  0.998 &  0.639 &  \bf{0.602} &  1.004 \\
B1-I33  &  1.115 &  0.696 &  0.698 &  0.723 &  \bf{0.672} &  0.722 &  0.711 \\
B1-I34  &  0.951 &  0.825 &  0.797 &  0.761 &  \bf{0.614} &  0.670 &  0.746 \\
B1-I35  &  0.853 &  0.917 &  0.901 &  0.883 &  \bf{0.779} &  0.833 &  0.846 \\
B1-I36  &  0.736 &  0.689 &  0.692 &  0.747 &  \bf{0.702} &  0.775 &  0.731 \\
B1-I37  &  1.325 &  1.349 &  1.397 &  1.334 &  \bf{0.529} &  0.647 &  1.343 \\
B1-I38  &  1.367 &  1.284 &  1.392 &  1.384 &  \bf{0.642} &  0.682 &  1.434 \\
B1-I39  &  0.897 &  0.874 &  0.885 &  0.894 &  \bf{0.585} &  0.598 &  0.873 \\
B1-I40  &  1.353 &  1.298 &  1.302 &  1.403 &  0.641 &  \bf{0.413} &  1.363 \\
B1-I41  &  0.858 &  0.803 &  0.769 &  0.770 &  \bf{0.635} &  0.734 &  0.749 \\
B1-I42  &  1.251 &  1.247 &  1.280 &  1.193 &  \bf{0.536} &  0.691 &  1.263 \\
B2-I1     &  0.958 &  1.163 &  1.107 &  0.848 &  \bf{0.761} &  0.852 &  0.955 \\
B2-I2     &  0.966 &  0.874 &  0.826 &  0.762 &  \bf{0.608} &  0.666 &  0.758 \\
B2-I3     &  0.829 &  0.836 &  0.825 &  0.723 &  \bf{0.569} &  0.700 &  0.724 \\
B2-I4     &  0.760 &  0.793 &  0.803 &  0.771 &  \bf{0.577} &  0.681 &  0.744 \\
B2-I5     &  0.836 &  0.780 &  0.768 &  0.737 &  \bf{0.537} &  0.699 &  0.733 \\
B2-I6     &  0.753 &  0.853 &  0.836 &  0.751 &  \bf{0.583} &  0.633 &  0.745 \\
B2-I7     &  0.853 &  0.805 &  0.813 &  0.760 &  0.696 &  \bf{0.635} &  0.747 \\
B2-I8     &  0.820 &  0.801 &  0.788 &  0.717 &  \bf{0.664} &  0.790 &  0.707 \\
B2-I9     &  0.830 &  0.703 &  0.698 &  0.716 &  0.639 &  \bf{0.637} &  0.718 \\
B2-I10    &  0.806 &  0.850 &  0.840 &  0.788 &  \bf{0.656} &  0.700 &  0.777 \\
B2-I11    &  0.831 &  0.806 &  0.800 &  0.826 &  \bf{0.673} &  0.788 &  0.826 \\
B2-I12    &  0.804 &  0.900 &  0.863 &  0.824 &  \bf{0.617} &  0.715 &  0.801 \\
B2-I13    &  0.834 &  0.806 &  0.793 &  0.816 &  \bf{0.604} &  0.631 &  0.823 \\
B2-I14    &  0.754 &  0.749 &  0.741 &  0.744 &  \bf{0.654} &  0.658 &  0.739 \\
B2-I15    &  0.686 &  0.785 &  0.760 &  0.739 &  \bf{0.604} &  0.679 &  0.685 \\
B2-I16    &  0.875 &  0.792 &  0.784 &  0.768 &  \bf{0.608} &  0.691 &  0.770 \\
B2-I17    &  0.984 &  0.888 &  0.860 &  0.790 &  \bf{0.665} &  0.696 &  0.792 \\
B2-I18    &  0.835 &  0.793 &  0.777 &  0.747 &  \bf{0.622} &  0.661 &  0.730 \\
B2-I19    &  1.351 &  1.153 &  1.144 &  0.986 &  \bf{0.781} &  0.833 &  1.062 \\
B2-I20    &  0.912 &  0.902 &  0.887 &  0.893 &  \bf{0.681} &  0.783 &  0.902 \\
B2-I21    &  0.951 &  0.774 &  0.759 &  0.734 &  \bf{0.602} &  0.650 &  0.740 \\
B2-I22    &  0.873 &  0.859 &  0.834 &  0.788 &  \bf{0.532} &  0.646 &  0.775 \\
B2-I23    &  0.849 &  0.816 &  0.795 &  0.768 &  \bf{0.577} &  0.643 &  0.752 \\
B2-I24    &  0.780 &  0.893 &  0.874 &  0.775 &  \bf{0.580} &  0.645 &  0.774 \\
B2-I25    &  0.846 &  0.709 &  0.702 &  0.732 &  \bf{0.659} &  0.721 &  0.735 \\
B2-I26    &  0.831 &  0.779 &  0.775 &  0.755 &  \bf{0.558} &  0.677 &  0.731 \\
B2-I27    &  0.819 &  0.817 &  0.797 &  0.724 &  \bf{0.548} &  0.614 &  0.720 \\
B2-I28    &  0.911 &  0.984 &  0.983 &  0.931 &  \bf{0.523} &  0.645 &  0.902 \\
B2-I29    &  0.985 &  0.804 &  0.806 &  0.795 &  \bf{0.505} &  0.689 &  0.785 \\
B2-I30    &  0.921 &  0.784 &  0.770 &  0.730 &  \bf{0.477} &  0.587 &  0.715 \\
B2-I31    &  0.761 &  0.808 &  0.789 &  0.767 &  \bf{0.564} &  0.629 &  0.759 \\
B2-I32    &  0.850 &  0.798 &  0.788 &  0.764 &  \bf{0.699} &  0.726 &  0.752 \\
B2-I33    &  0.800 &  0.737 &  0.746 &  0.726 &  \bf{0.515} &  0.644 &  0.699 \\
B2-I34    &  0.712 &  0.781 &  0.751 &  0.689 &  \bf{0.511} &  0.650 &  0.687 \\
B2-I35    &  0.865 &  0.791 &  0.778 &  0.766 &  0.614 &  \bf{0.570} &  0.754 \\
B2-I36    &  0.808 &  0.735 &  0.718 &  0.694 &  0.674 &  \bf{0.604} &  0.691 \\
B2-I37    &  0.777 &  0.823 &  0.818 &  0.806 &  0.676 &  \bf{0.670} &  0.771 \\
B2-I38    &  0.728 &  0.727 &  0.723 &  0.732 &  0.633 &  \bf{0.628} &  0.714 \\
B2-I39    &  0.780 &  0.994 &  0.946 &  0.771 &  0.670 &  \bf{0.609} &  0.764 \\
B2-I40    &  0.825 &  0.731 &  0.728 &  0.702 &  \bf{0.667} &  0.704 &  0.711 \\
B2-I41    &  0.856 &  0.743 &  0.740 &  0.759 &  \bf{0.635} &  0.793 &  0.758 \\
B2-I42    &  0.853 &  0.773 &  0.754 &  0.729 &  \bf{0.689} &  0.771 &  0.738 \\
B2-I43    &  0.762 &  0.801 &  0.794 &  0.749 &  \bf{0.580} &  0.777 &  0.747 \\
B2-I44    &  0.945 &  0.769 &  0.762 &  0.732 &  \bf{0.586} &  0.655 &  0.729 \\
B2-I45    &  0.966 &  0.890 &  0.871 &  \bf{0.763} &  0.774 &  0.816 &  0.866 \\
B3-I1  &  1.045 &  1.358 &  1.322 &  1.089 &  \bf{0.723} &  0.759 &  1.058 \\
B3-I2  &  1.021 &  1.196 &  1.256 &  1.131 &  \bf{0.575} &  0.794 &  1.041 \\
B3-I3  &  0.990 &  0.881 &  0.910 &  1.160 &  \bf{0.619} &  0.660 &  0.998 \\
B3-I4  &  0.953 &  0.939 &  0.943 &  0.906 &  \bf{0.686} &  0.719 &  0.961 \\
B3-I5  &  0.907 &  0.991 &  0.993 &  1.032 &  \bf{0.521} &  0.659 &  0.906 \\
B3-I6  &  1.116 &  0.981 &  0.965 &  1.067 &  \bf{0.653} &  0.729 &  1.019 \\
B3-I7  &  1.223 &  0.905 &  0.904 &  0.859 &  \bf{0.649} &  0.787 &  0.914 \\
B3-I8  &  1.374 &  1.204 &  1.267 &  0.974 &  \bf{0.526} &  0.679 &  1.122 \\
B3-I9  &  1.382 &  1.123 &  1.130 &  0.917 &  \bf{0.655} &  0.661 &  1.176 \\
B3-I10 &  1.357 &  1.408 &  1.483 &  1.189 &  0.681 &  {0.611} &  1.111 \\
B3-I11 &  1.014 &  1.197 &  1.217 &  1.032 &  0.784 &  \bf{0.708} &  1.022 \\
B3-I12 &  0.934 &  0.822 &  0.815 &  0.901 &  \bf{0.621} &  0.722 &  0.834 \\
B3-I13 &  0.918 &  0.903 &  0.892 &  0.958 &  \bf{0.738} &  0.839 &  0.915 \\
B3-I14 &  1.230 &  1.351 &  1.369 &  1.342 &  \bf{0.788} &  0.606 &  1.232 \\
B3-I15 &  0.928 &  0.976 &  0.986 &  0.935 &  \bf{0.679} &  0.760 &  0.925 \\
B3-I16 &  1.162 &  1.324 &  1.390 &  0.957 &  \bf{0.644} &  0.856 &  1.090 \\
B3-I17 &  0.993 &  1.153 &  1.206 &  1.016 &  \bf{0.634} &  0.862 &  1.006 \\
B3-I18 &  0.918 &  1.118 &  1.088 &  0.895 &  \bf{0.675} &  0.775 &  0.927 \\
B3-I19 &  0.870 &  0.879 &  0.871 &  0.842 &  \bf{0.620} &  0.673 &  0.877 \\
B3-I20 &  0.882 &  0.913 &  0.902 &  0.894 &  \bf{0.677} &  0.824 &  0.870 \\
B3-I21 &  1.060 &  1.232 &  1.276 &  1.199 &  \bf{0.633} &  0.653 &  1.017 \\
B4-I1     &  1.287 &  1.094 &  1.161 &  1.176 &  \bf{0.561} &  0.766 &  1.186 \\
B4-I2     &  1.242 &  1.780 &  1.971 &  1.130 &  \bf{0.620} &  0.666 &  1.077 \\
B4-I3     &  1.041 &  0.889 &  0.913 &  0.914 &  0.718 &  \bf{0.670} &  0.938 \\
B4-I4     &  1.200 &  1.055 &  1.125 &  1.056 &  \bf{0.630} &  0.710 &  0.968 \\
B4-I5     &  1.322 &  1.245 &  1.341 &  1.094 &  0.622 &  \bf{0.607} &  1.318 \\
B4-I6     &  1.478 &  1.543 &  1.659 &  1.168 &  0.684 &  \bf{0.654} &  0.886 \\
B4-I7     &  1.324 &  0.993 &  1.044 &  0.822 &  0.647 &  \bf{0.624} &  0.919 \\
B4-I8     &  1.275 &  1.106 &  1.122 &  1.083 &  \bf{0.634} &  0.704 &  0.961 \\
B4-I9     &  1.107 &  1.399 &  1.528 &  0.837 &  0.854 &  0.861 &  \bf{0.830} \\
B4-I10    &  1.306 &  1.334 &  1.469 &  1.083 &  0.769 &  \bf{0.766} &  0.804 \\
\midrule
Average Item & 0.981 &	0.949 &	0.956	& 0.909 &	\bf{0.644} &	0.697 &	0.875 \\
\bottomrule
\label{tab:results_pasta}
\end{tabularx}

\begin{figure}
  \centering
  \includegraphics[scale=0.8]{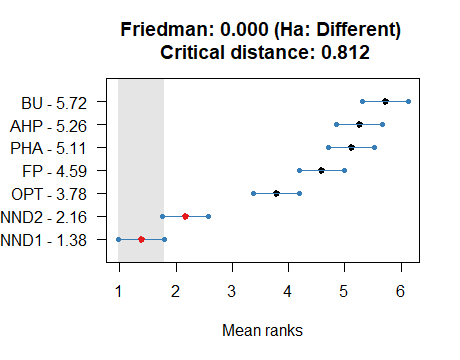}
 \caption{Nemenyi test results at 95\% confidence level for all 123 series of the Italian sales dataset. The hierarchical forecasting methods are sorted vertically according to the MASE mean rank.}
 \label{fig:fried_italian}
\end{figure}

In Figure \ref{fig:fried_italian} we show the outcome of the Friedman and post-hoc Nemenyi tests with a confidence level of 95\%.  If the intervals of two methods do not overlap,
they exhibit statistically different performance. As seen, Table \ref{tab:results_pasta} and Figure \ref{fig:fried_italian} indicate that NND (iterative and standard top-down) provide significantly better forecasts than the rest of the methods found in the literature, with the bottom-up performing worst. The bad performance of the bottom-up method can be attributed to the demand at the most granular level of the hierarchy being challenging to model and forecast effectively due to its too sparse and erratic nature. The majority of the item-level time series display sporadic sales including zeros, and the promotion of an item does not always correspond to an increase in sales. By using traditional or combination of methods to generate base forecasts for the time series at the lowest level, we end up with flat line forecasts, representing the average demand, failing to account for the seasonality that truly exists but is impossible to identify between the noise. By focusing our attention at the highest or some intermediate level of the hierarchy, we have enough data to build decent models capturing the underlying trend and  seasonality.Indeed, the aggregation tends to regularize the demand and make it easier to forecast. The only level for which the optimal reconciliation approach is the best is the top level. As we move down the hierarchy our approach outperforms all the top-down approaches, the bottom-up method and the optimal reconciliation, with the NND iterative top-down (NND2) performing best at the brand level and the NND standard top-down (NND1) performing best at the item level, on average.

\begin{figure}[!ht]
  \centering
  \includegraphics[scale=0.8]{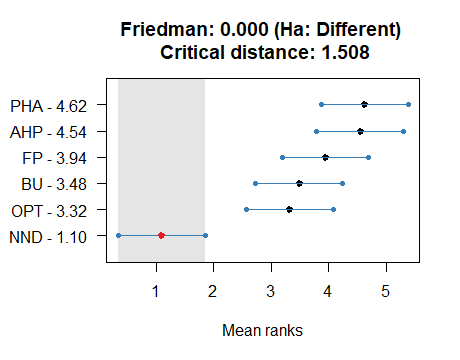}
 \caption{Nemenyi test results at 95\% confidence level for all 25 series of the Electricity data. The hierarchical forecasting methods are sorted vertically according to the MASE mean rank.}
 \label{fig:fried_elec_mase}
\end{figure}

\begin{figure}
  \centering
  \includegraphics[scale=0.8]{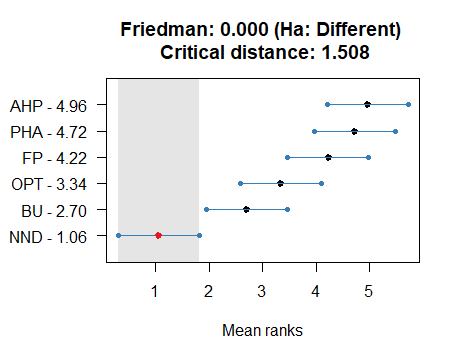}
 \caption{Nemenyi test results at 95\% confidence level for all 25 series of the Electricity data. The hierarchical forecasting methods are sorted vertically according to the SMAPE mean rank.}
 \label{fig:fried_elec_smape}
\end{figure}

In Table \ref{tab:results_electricity_mase} and \ref{tab:results_electricity_smape} we present the MASE and SMAPE for all 25 time series of the electricity demand dataset. Note that here we only have two levels, so that NND1 and NND2 coincide (which is why we call it only NND in the tables).
In Figure \ref{fig:fried_elec_mase} and \ref{fig:fried_elec_smape} we show the outcome of the Friedman and Nemenyi  tests with a confidence level of 95\%, with respect to MASE and SMAPE.
We find that all the top-down approaches perform best at the grid level. On this dataset, the optimal reconciliation method, and the bottom-up approach show good performance. The good performance of the bottom-up method with respect to the classical top-down approaches can be attributed to the strong seasonality of the series, even at the bottom level. Our NND clearly outperforms all the competitors, ranking first in the tests and having a better average error.

\begin{tabularx}{\textwidth}{lcccccc}
\caption{MASE for all 25 series of the electricity demand data. In bold the best performing approach.} \\
\toprule
{} &     BU &    AHP &    PHA &     FP &    NND &    OPT \\
\midrule
TOTAL    &  1.090 &  \bf{0.974} &  \bf{0.974} &  \bf{0.974} &  \bf{0.974} &  1.088 \\
\midrule
M1  &  1.745 &  1.520 &  1.549 &  1.092 &  \bf{1.002} &  1.249 \\
M2  &  1.759 &  1.062 &  1.271 &  1.174 &  \bf{1.013} &  1.432 \\
M3  &  1.158 &  1.361 &  1.355 &  0.963 &  \bf{0.892} &  1.170 \\
M4  &  1.475 &  1.612 &  1.694 &  1.128 &  \bf{1.038} &  1.465 \\
M5  &  0.999 &  1.075 &  1.063 &  1.181 &  \bf{0.873} &  1.132 \\
M6  &  1.352 &  1.486 &  1.520 &  1.821 &  \bf{0.769} &  1.621 \\
M7  &  1.044 &  1.771 &  1.678 &  1.174 &  \bf{0.637} &  1.087 \\
M8  &  1.029 &  1.042 &  1.041 &  1.323 &  \bf{0.892} &  1.053 \\
M9  &  1.070 &  1.272 &  1.244 &  0.997 &  \bf{0.893} &  0.975 \\
M10 &  \bf{1.045} &  1.573 &  1.564 &  1.620 &  1.051 &  1.340 \\
M11 &  1.225 &  1.452 &  1.437 &  1.382 &  \bf{0.840} &  1.131 \\
M12 &  1.430 &  1.468 &  1.411 &  1.163 &  \bf{0.821} &  1.342 \\
M13 &  \bf{1.002} &  1.872 &  1.715 &  1.496 &  1.090 &  1.682 \\
M14 &  1.041 &  1.162 &  1.164 &  1.391 &  \bf{0.820} &  1.121 \\
M15 &  1.348 &  1.533 &  1.553 &  1.893 &  \bf{0.863} &  1.185 \\
M16 &  1.095 &  1.055 &  1.174 &  1.831 &  \bf{0.844} &  1.457 \\
M17 &  1.292 &  1.806 &  1.759 &  1.706 &  \bf{1.122} &  1.131 \\
M18 &  1.263 &  1.824 &  1.826 &  1.393 &  \bf{1.047} &  1.299 \\
M19 &  1.436 &  1.730 &  1.789 &  1.218 &  \bf{0.882} &  1.280 \\
M20 &  1.240 &  1.310 &  1.331 &  1.008 &  \bf{0.987} &  1.016 \\
M21 &  1.326 &  1.317 &  1.343 &  1.176 &  \bf{0.802} &  1.106 \\
M22 &  1.431 &  1.108 &  1.078 &  1.316 &  \bf{0.924} &  1.380 \\
M23 &  1.164 &  1.353 &  1.309 &  1.419 &  0\bf{.830} &  1.153 \\
M24 &  1.284 &  1.705 &  1.740 &  1.532 &  \bf{1.165} &  1.232 \\
\midrule
Average Meter & 1.261 &	1.436 &	1.442 &	1.350 &	\bf{0.921} & 1.252 \\
\bottomrule
\label{tab:results_electricity_mase}
\end{tabularx}

\begin{tabularx}{\textwidth}{lcccccc}
\caption{SMAPE for all 25 series of the electricity demand data. In bold the best performing approach.} \\
\toprule
{} &     BU &    AHP &    PHA &     FP &    NND &    OPT \\
\midrule
TOTAL    &  0.076 &  \bf{0.072} &  \bf{0.072} &  \bf{0.072} &  \bf{0.072} &  0.075 \\
\midrule
M1  &  0.270 &  0.229 &  0.233 &  0.235 &  \bf{0.165} &  0.267 \\
M2  &  0.233 &  0.272 &  0.257 &  0.201 &  \bf{0.099} &  0.246 \\
M3  &  0.218 &  0.231 &  0.230 &  0.244 &  \bf{0.136} &  0.248 \\
M4  &  0.183 &  0.296 &  0.393 &  0.212 &  \bf{0.131} &  0.224 \\
M5  &  0.217 &  0.338 &  0.336 &  0.278 &  \bf{0.118} &  0.268 \\
M6  &  0.195 &  0.274 &  0.279 &  0.294 &  \bf{0.072} &  0.192 \\
M7  &  0.214 &  0.305 &  0.308 &  0.351 &  \bf{0.176} &  0.213 \\
M8  &  0.228 &  0.366 &  0.265 &  0.238 &  \bf{0.155} &  0.247 \\
M9  &  0.217 &  0.294 &  0.287 &  0.216 &  \bf{0.112} &  0.213 \\
M10 &  0.216 &  0.259 &  0.258 &  0.274 &  \bf{0.146} &  0.226 \\
M11 &  0.237 &  0.313 &  0.292 &  0.246 &  \bf{0.144} &  0.237 \\
M12 &  0.211 &  0.219 &  0.280 &  0.276 &  \bf{0.089} &  0.206 \\
M13 &  0.222 &  0.381 &  0.374 &  0.294 &  \bf{0.095} &  0.225 \\
M14 &  0.195 &  0.318 &  0.318 &  0.260 &  \bf{0.104} &  0.227 \\
M15 &  0.205 &  0.388 &  0.322 &  0.273 &  \bf{0.121} &  0.271 \\
M16 &  0.198 &  0.235 &  0.246 &  0.108 &  \bf{0.063} &  0.272 \\
M17 &  0.193 &  0.235 &  0.228 &  0.295 &  \bf{0.124} &  0.199 \\
M18 &  0.209 &  0.271 &  0.272 &  0.256 &  \bf{0.164} &  0.224 \\
M19 &  0.194 &  0.292 &  0.206 &  0.226 &  \bf{0.134} &  0.251 \\
M20 &  0.207 &  0.320 &  0.308 &  0.340 &  \bf{0.106} &  0.244 \\
M21 &  0.192 &  0.301 &  0.307 &  0.262 &  \bf{0.128} &  0.192 \\
M22 &  0.183 &  0.225 &  0.220 &  0.302 &  \bf{0.112} &  0.132 \\
M23 &  0.185 &  0.390 &  0.381 &  0.208 &  \bf{0.141} &  0.197 \\
M24 &  0.198 &  0.318 &  0.325 &  0.309 &  \bf{0.145} &  0.188 \\
\midrule
Average Meter & 0.209 &	0.294 &	0.288 &	0.258 &	\bf{0.124} &	0.225 \\
\bottomrule
\label{tab:results_electricity_smape}
\end{tabularx}

In Table \ref{tab:results_m5}, we provide results for all 44 time series of the Walmart dataset.
For the NND1, we directly generate forecasts at the category level using the total aggregate at level 0. We aggregate these forecasts to obtain first the store-level forecasts, and then the state-level forecasts. For the NND2, instead, we train a disaggregation model that outputs the state-level forecasts starting from the total aggregate series at level 0, one NND for each state-level series to generate forecasts for the stores of the geographical area they belong to, and finally, one model for each store to obtain the bottom-level forecast at the category level. Overall, for the entire hierarchy, we train one NND at the top level, 3 NND in parallel at the state level, and 10 NND in parallel at the store level.

\newpage

\begin{tabularx}{\textwidth}{lccccccc}
\caption{MASE for all 44 series of the Walmart data. In bold the best performing approach.} \\
\toprule
{} &     BU &    AHP &    PHA &     FP &   NND1 &   NND2 &    OPT \\
\midrule
TOTAL      &  \bf{0.781} &  0.782 &  0.782 &  0.782 &  0.782 &  0.782 &  0.785 \\
\midrule
CA         &  0.807 &  0.997 &  0.997 &  0.877 &  0.593 &  \bf{0.571} &  0.810 \\
TX         &  0.731 &  0.986 &  0.956 &  0.751 &  0.632 &  \bf{0.615} &  0.734 \\
WI         &  0.760 &  0.858 &  0.846 &  0.778 &  0.578 &  \bf{0.492} &  0.760 \\
\midrule
Average State & 0.766 & 0.947	& 0.933 & 0.802 & 0.601 & \bf{0.559} & 0.768 \\
\midrule
CA1     &  0.771 &  1.044 &  1.070 &  0.828 &  \bf{0.687} &  0.761 &  0.775 \\
CA2     &  0.843 &  1.029 &  1.034 &  0.887 &  \bf{0.625} &  0.669 &  0.839 \\
CA3     &  0.876 &  1.371 &  1.340 &  0.914 &  0.576 &  \bf{0.494} &  0.873 \\
CA4     &  0.757 &  1.023 &  1.020 &  0.800 &  0.593 &  \bf{0.576} &  0.758 \\
TX1     &  0.739 &  0.968 &  0.965 &  0.762 &  0.572 &  \bf{0.446} &  0.743 \\
TX2     &  0.756 &  1.518 &  1.451 &  0.761 &  0.651 &  \bf{0.579} &  0.755 \\
TX3     &  0.723 &  0.761 &  0.764 &  0.742 &  0.577 &  \bf{0.559} &  0.723 \\
WI1     &  0.746 &  1.454 &  1.369 &  0.745 &  0.796 &  \bf{0.741} &  0.743 \\
WI2     &  0.737 &  0.907 &  0.888 &  0.754 &  0.561 &  \bf{0.572} &  0.740 \\
WI3     &  0.813 &  1.017 &  0.987 &  0.845 &  0.516 &  \bf{0.483} &  0.818 \\
\midrule
Average Store & 0.776 &	1.109 &	1.089 &	0.804 &	0.615 & \bf{0.588} & 0.777 \\
\midrule
CA1-Foods &  0.823 &  1.385 &  1.432 &  0.864 &  \bf{0.521} &  0.590 &  0.827 \\
CA1-Hobbies &  0.722 &  0.840 &  0.848 &  0.735 &  0.733 &  \bf{0.686} &  0.722 \\
CA1-Household &  0.742 &  1.328 &  1.628 &  0.798 &  \bf{0.673} &  0.689 &  0.743 \\
CA2-Food &  0.808 &  1.060 &  1.111 &  0.838 &  \bf{0.711} &  0.713 &  0.809 \\
CA2-Hobbies &  0.792 &  1.015 &  1.020 &  0.819 &  \bf{0.753} &  0.755 &  0.791 \\
CA2-Household &  0.800 &  1.549 &  1.507 &  0.826 &  0.783 &  \bf{0.781} &  0.796 \\
CA3-Food &  0.864 &  1.393 &  1.317 &  0.887 &  \bf{0.754} &  0.788 &  0.862 \\
CA3-Hobbies &  0.749 &  0.889 &  0.898 &  0.775 &  0.697 &  \bf{0.647} &  0.750 \\
CA3-Household &  0.807 &  1.283 &  1.256 &  0.884 &  \bf{0.745} &  0.777 &  0.811 \\
CA4-Food &  0.770 &  1.199 &  1.161 &  0.803 &  \bf{0.726} &  0.747 &  0.771 \\
CA4-Hobbies &  0.709 &  0.967 &  0.980 &  0.720 &  0.709 &  \bf{0.667} &  0.709 \\
CA4-Household &  0.712 &  1.402 &  1.389 &  0.758 &  0.692 &  \bf{0.631} &  0.713 \\
TX1-Food &  0.761 &  1.347 &  1.315 &  0.773 &  0.644 &  \bf{0.523} &  0.763 \\
TX1-Hobbies &  0.728 &  0.912 &  0.914 &  0.738 &  0.756 &  \bf{0.719} &  0.728 \\
TX1-Household &  0.770 &  1.218 &  1.164 &  0.808 &  0.679 &  \bf{0.660} &  0.771 \\
TX2-Food &  0.778 &  1.308 &  1.800 &  0.774 &  \bf{0.615} &  0.699 &  0.776 \\
TX2-Hobbies &  0.722 &  0.852 &  0.848 &  0.724 &  0.773 &  \bf{0.622} &  0.721 \\
TX2-Household &  0.772 &  0.974 &  0.965 &  0.794 &  0.652 &  \bf{0.628} &  0.771 \\
TX3-Food &  0.735 &  0.845 &  0.813 &  0.745 &  0.655 &  \bf{0.582} &  0.735 \\
TX3-Hobbies &  0.739 &  1.334 &  1.329 &  0.749 &  \bf{0.646} &  0.679 &  0.739 \\
TX3-Household &  0.771 &  1.193 &  1.143 &  0.793 &  \bf{0.714} &  0.776 &  0.769 \\
WI1-Food &  0.760 &  1.489 &  1.374 &  0.755 &  \bf{0.622} &  0.711 &  0.756 \\
WI1-Hobbies &  0.708 &  0.992 &  0.992 &  0.710 &  \bf{0.604} &  0.622 &  0.708 \\
WI1-Household &  0.761 &  1.361 &  1.316 &  0.766 &  \bf{0.741} &  0.779 &  0.760 \\
WI2-Food &  0.727 &  0.771 &  0.767 &  0.732 &  0.657 &  \bf{0.566} &  0.728 \\
WI2-Hobbies &  0.749 &  1.057 &  1.052 &  0.767 &  \bf{0.634} &  0.671 &  0.749 \\
WI2-Household &  0.847 &  1.173 &  1.123 &  0.883 &  \bf{0.775} &  0.681 &  0.849 \\
WI3-Food &  0.779 &  1.031 &  0.996 &  0.797 &  0.709 &  \bf{0.605} &  0.780 \\
WI3-Hobbies &  0.748 &  0.924 &  0.897 &  0.763 &  \bf{0.605} &  0.663 &  0.748 \\
WI3-Household &  0.825 &  0.878 &  0.874 &  0.873 &  0.737 &  \bf{0.710} &  0.826 \\
\midrule
Average Category & 0.765 &	1.132 & 1.141 &	0.788 &	0.690 & \bf{0.679} & 0.768 \\
\bottomrule
\label{tab:results_m5}
\end{tabularx}

\begin{figure}
  \centering
  \includegraphics[scale=0.8]{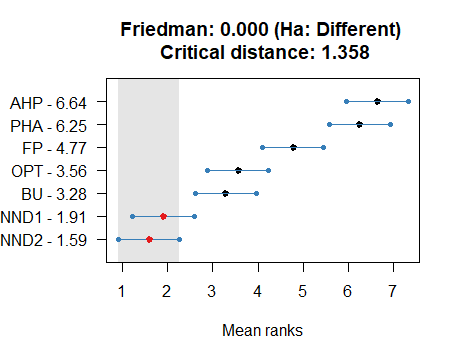}
 \caption{Nemenyi test results at 95\% confidence level for all 44 series of the Walmart data. The hierarchical forecasting methods are sorted vertically according to the MASE mean rank.}
 \label{fig:fried_m5}
\end{figure}

\newpage

In Figure \ref{fig:fried_m5}, we plot the results of the Friedman and Nemenyi tests for all 44 series of the Walmart dataset. Figure \ref{fig:fried_m5} shows that NND1 and NND2 are statistically equivalent on this dataset, even though, on average, the error produced by NND2 is lower. Anyhow, both NND1 and NND2  outperform the competitors. On this dataset, the bottom-up approach performs best at the most aggregate level, and it is quite competitive with the optimal combination approach since the time series at the category level display a strong seasonality component. Indeed, sales are relatively high on weekends in comparison to normal days, and this behavior propagates as we go up the hierarchy. As we move down the hierarchy, our approach outperforms all the top-down approaches, the bottom-up method, and the optimal reconciliation.

Summarizing, our method outperforms the hierarchical forecasting competitors on all the considered datasets. This result is particularly significant due to the different characteristics of the three datasets. In the sales data, the bottom-level series are extremely noisy and hard to forecast (as confirmed by the bottom-up method's poor performance). On the other hand, the electricity demand data display seasonality at the bottom
 level, as confirmed by the bottom-up method's good performance. Finally, the Walmart dataset comes from a domain similar to the one of the Italian Sales Data, has a higher number of levels, but a strong seasonality that propagates through the hierarchy even at the bottom level.
In all the experiments, our approach generates coherent forecasts: the maximum violation of the aggregation constraint is less than $10^{-3}$. Furthermore, it improves the overall accuracy at any level of the hierarchy. This confirms the general viability of our approach, which can get coherent and accurate forecasts by extracting the hidden information in the hierarchy.\par
In \ref{sec:app2}, we also report some figures to show the accuracy of the forecasts produced by our method on some series of the considered datasets at different levels of the hierarchy.

\section{Conclusions}
In this paper, we propose a machine learning method for forecasting hierarchical time series. Our approach relies on a deep neural network capable of automatically extracting time series features at any level thanks to the convolutional layers. The network combines these features with the explanatory variables available at any level of the hierarchy. The obtained forecasts are coherent since reconciliation is forced in the training phase by minimizing a customized loss function. The effectiveness of the approach is shown on three real-world datasets that fit the method's assumptions: they include explanatory variables, and the number of observations is large enough. On these datasets, a deep statistical analysis proves that our method outperforms the state-of-the-art competitors in hierarchical forecasting, being always more accurate, and producing significantly different forecasts at any level. The results confirm that we fulfilled our aim to combine in a single machinery all the available information through the hierarchy, both hidden and provided by the explanatory variables, without the need of post-processing on the series to achieve high accuracy and cross-sectional coherence.\par
As future work, our idea is to extend the proposed methodology to take into account the temporal reconciliation and jointly perform the temporal and cross-sectional reconciliation. Forcing only the temporal coherence can be viewed as a straightforward extension, whereas keeping into account both temporal and cross-sectional coherence may require some changes to the network structure. The idea is to adapt the neural network to exploit the information for the temporal reconciliation, which may require some recurrent layers, as LSTM or GRU. Also, the loss function should be adapted to effectively force both reconciliation constraints during the training.

\section*{Acknowledgments}
We would like to thank the reviewers for their thoughtful comments that greatly helped to improve our manuscript.

\appendix
\section{Details on the NND hyperparameters}
\label{sec:app1}

In Table \ref{tab:details-NND1}, we report for each dataset the details of the implemented neural network used for producing the results of method NND standard top-down (NND1) described in Section \ref{sec:nnd}. We have one row for each dataset, whereas on the columns we have the level of disaggregation, the number of units in all the dense layers, the number of filters and kernel size of all the convolutional layers.

\begin{table}[ht]
    \centering
    \begin{tabular}{|l|l|c|c|c|}\hline
         Dataset & Level & units dense &  filters & kernel size  \\\hline
         Italian & Total to Item & 128 & 16 & 8\\\hline\hline
         Electricity  & Total to  Meter & 256 & 32 & 16\\\hline\hline
         Walmart  & Total to Category & 128 & 32 & 4\\\hline        
    \end{tabular}
    \caption{Implementation details for the networks used in NND1}
    \label{tab:details-NND1}
\end{table}

In Table \ref{tab:details-NND2}, we report for each dataset the details of the implemented neural networks used for producing the results of the method NND iterative top-down (NND2) described in Section \ref{sec:nnd}. We have one row for each dataset and for each network used in NND2 for that dataset, whereas on the columns we have the level of disaggregation, the number of units in all the dense layers, the number of filters and kernel size of all the convolutional layers. Depending on the dataset, the number of rows changes depending on the number of models built for a given level.

\begin{table}[ht]
    \centering
    \begin{tabular}{|l|l|c|c|c|}\hline
        Dataset & Level & units dense & filters & kernel size  \\\hline
         Italian & Total to Brand & 64 & 32 & 8\\\hline 
         Italian &  B1 to B1-I$X$ &  128 & 16 & 4\\\hline 
         Italian &  B2 to B2-I$X$ & 64 & 32 & 4\\\hline 
         Italian &  B3 to B3-I$X$ & 64 & 16 & 4\\\hline 
         Italian &  B4 to B4-I$X$ & 128 & 16 & 4\\\hline\hline
         Walmart  & Total to State & 128 & 16  & 8 \\\hline
         Walmart  & CA to CA$X$ & 64 & 16 & 4 \\\hline
         Walmart  & TX to TX$X$ & 64 & 16 & 8 \\\hline
         Walmart  & WI to WI$X$ & 128 & 16 & 8\\\hline
         Walmart  & CA1 to CA1-Category & 64 & 32 & 8\\\hline
         Walmart  & CA2 to CA2-Category & 64 & 16 & 8\\\hline
         Walmart  & CA3 to CA3-Category & 64 & 16 & 8\\\hline
         Walmart  & CA4 to CA4-Category & 64 & 16 & 4\\\hline
         Walmart  & TX1 to TX1-Category & 64 & 16 & 8\\\hline
         Walmart  & TX2 to TX2-Category & 64 & 32 & 8\\\hline
         Walmart  & TX3 to TX3-Category & 128 & 32 & 4\\\hline
         Walmart  & WI1 to WI1-Category & 64 & 16 & 8\\\hline
         Walmart  & WI2 to WI2-Category & 128 & 16 & 8\\\hline
         Walmart  & WI3 to WI3-Category & 64 & 32 & 8\\\hline
    \end{tabular}
    \caption{Implementation details for the networks used in NND2}
    \label{tab:details-NND2}
\end{table}

\section{Plots}
\label{sec:app2}

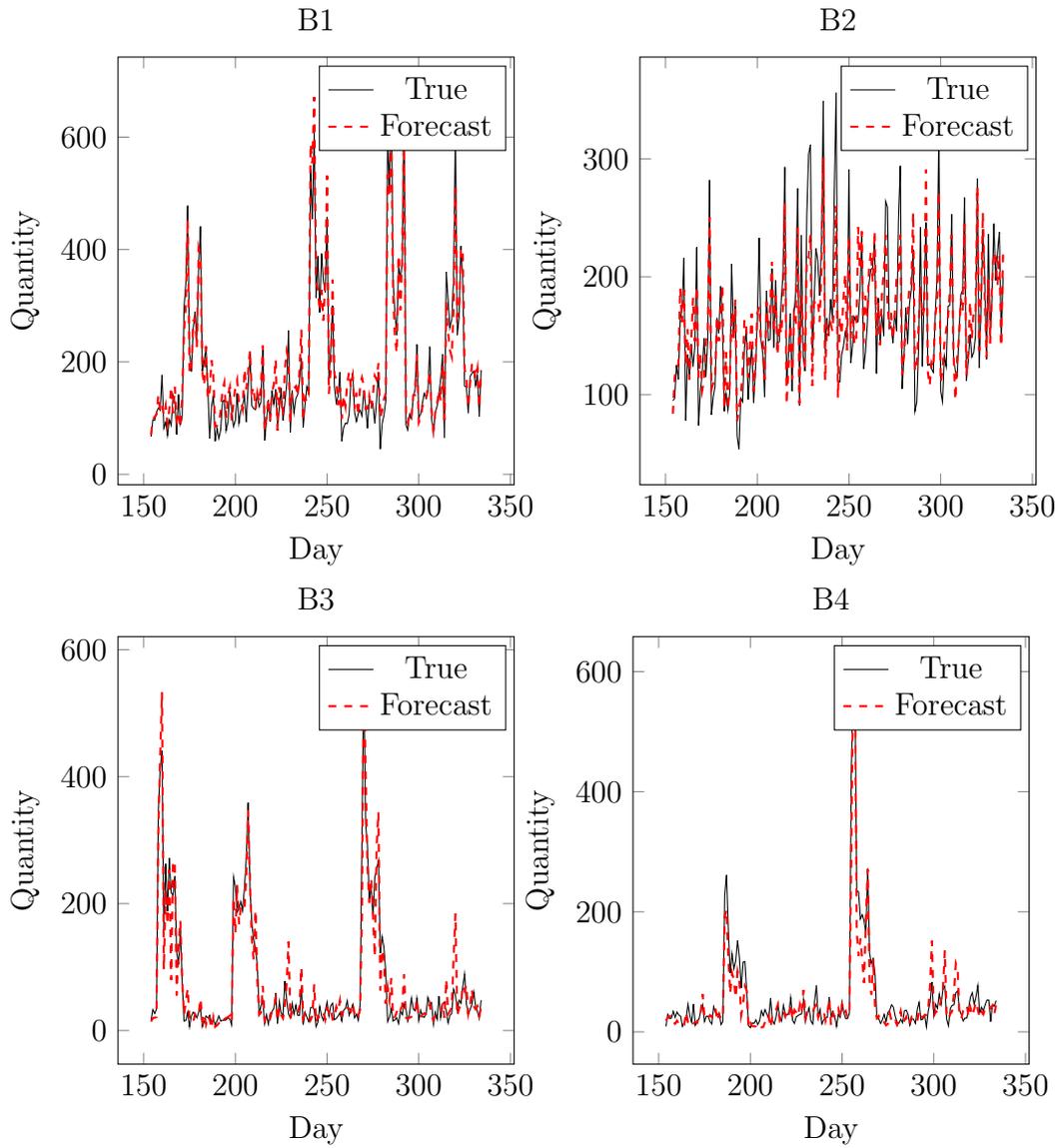
\begin{figure}

\begin{subfigure}{0.5\linewidth}
 
    \begin{tikzpicture}
      \begin{axis}[
      title={B1},
      xlabel={Day},
      ylabel={Quantity}, 
      width=\textwidth,
      height=\axisdefaultheight]
      
        \addplot[black] table[x=day,y=test_barilla,col sep=comma]{brand_table.csv};
        \addplot[thick, red, dashed] table[x=day,y=pred_barilla,col sep=comma]{brand_table.csv};
        \legend{True, Forecast}
        
      \end{axis}
      
    \end{tikzpicture}
 
\end{subfigure} %
\begin{subfigure}{0.5\linewidth}
 
    \begin{tikzpicture}
      \begin{axis}[
      title={B2},
      xlabel={Day},
      ylabel={Quantity},
      width=\textwidth,
      height=\axisdefaultheight]
      
        \addplot[black] table[x=day,y=test_conad,col sep=comma]{brand_table.csv};
        \addplot[thick, red, dashed] table[x=day,y=pred_conad,col sep=comma]{brand_table.csv};
        \legend{True, Forecast}
        
      \end{axis}
      
    \end{tikzpicture}
 
\end{subfigure}

\begin{subfigure}{0.5\textwidth}
 
    \begin{tikzpicture}
      \begin{axis}[
      title={B3},
      xlabel={Day},
      ylabel={Quantity},
      width=\textwidth,
      height=\axisdefaultheight]
      
        \addplot[black] table[x=day,y=test_cecco,col sep=comma]{brand_table.csv};
        \addplot[thick, red, dashed] table[x=day,y=pred_cecco,col sep=comma]{brand_table.csv};
        \legend{True, Forecast}
        
      \end{axis}
      
    \end{tikzpicture}
 
\end{subfigure}%
\begin{subfigure}{0.5\textwidth}
 
    \begin{tikzpicture}
      \begin{axis}[
      title={B4},
      xlabel={Day},
      ylabel={Quantity},
      width=\textwidth,
      height=\axisdefaultheight]
      
        \addplot[black] table[x=day,y=test_garofalo,col sep=comma]{brand_table.csv};
        \addplot[thick, red, dashed] table[x=day,y=pred_garofalo,col sep=comma]{brand_table.csv};
        \legend{True, Forecast}
        
      \end{axis}
      
    \end{tikzpicture}
 
\end{subfigure}

\caption{NND out-of-sample forecasts for the Italian sales dataset. Series B1, B2, B3 and B4 (last 6 months).}
\label{fig:italian_plot}
\end{figure}

\begin{figure}

\begin{subfigure}{0.5\linewidth}
 
    \begin{tikzpicture}
      \begin{axis}[
      title={CA4-Household},
      xlabel={Day},
      ylabel={Quantity}, 
      width=\textwidth,
      height=\axisdefaultheight]
      
        \addplot[black] table[x=day,y=true_CA4-Household,col sep=comma]{walmart_table.csv};
        \addplot[thick, red, dashed] table[x=day,y=pred_CA4-Household,col sep=comma]{walmart_table.csv};
        \legend{True, Forecast}
        
      \end{axis}
      
    \end{tikzpicture}
 
\end{subfigure} %
\begin{subfigure}{0.5\linewidth}
 
    \begin{tikzpicture}
      \begin{axis}[
      title={CA1-Foods},
      xlabel={Day},
      ylabel={Quantity},
      width=\textwidth,
      height=\axisdefaultheight]
      
        \addplot[black] table[x=day,y=true_CA1-Foods,col sep=comma]{walmart_table.csv};
        \addplot[thick, red, dashed] table[x=day,y=pred_CA1-Foods,col sep=comma]{walmart_table.csv};
        \legend{True, Forecast}
        
      \end{axis}
      
    \end{tikzpicture}
 
\end{subfigure}

\begin{subfigure}{0.48\textwidth}
 
    \begin{tikzpicture}
      \begin{axis}[
      title={TX1},
      xlabel={Day},
      ylabel={Quantity},
      width=\textwidth,
      height=\axisdefaultheight]
      
        \addplot[black] table[x=day,y=true_TX1,col sep=comma]{walmart_table.csv};
        \addplot[thick, red, dashed] table[x=day,y=pred_TX1,col sep=comma]{walmart_table.csv};
        \legend{True, Forecast}
        
      \end{axis}
      
    \end{tikzpicture}
 
\end{subfigure}\hspace{4mm}%
\begin{subfigure}{0.48\textwidth}
 
    \begin{tikzpicture}
      \begin{axis}[
      title={WI2-Foods},
      xlabel={Day},
      ylabel={Quantity},
      width=\textwidth,
      height=\axisdefaultheight]
      
        \addplot[black] table[x=day,y=true_WI2-Foods,col sep=comma]{walmart_table.csv};
        \addplot[thick, red, dashed] table[x=day,y=pred_WI2-Foods,col sep=comma]{walmart_table.csv};
        \legend{True, Forecast}
        
      \end{axis}
      
    \end{tikzpicture}
 
\end{subfigure}

\caption{NND out-of-sample forecasts for the Walmart dataset. Time series CA4-Household, CA1-Foods, TX1 and WI2-Foods (last 6 months).}
\label{fig:walmart_plot}
\end{figure}
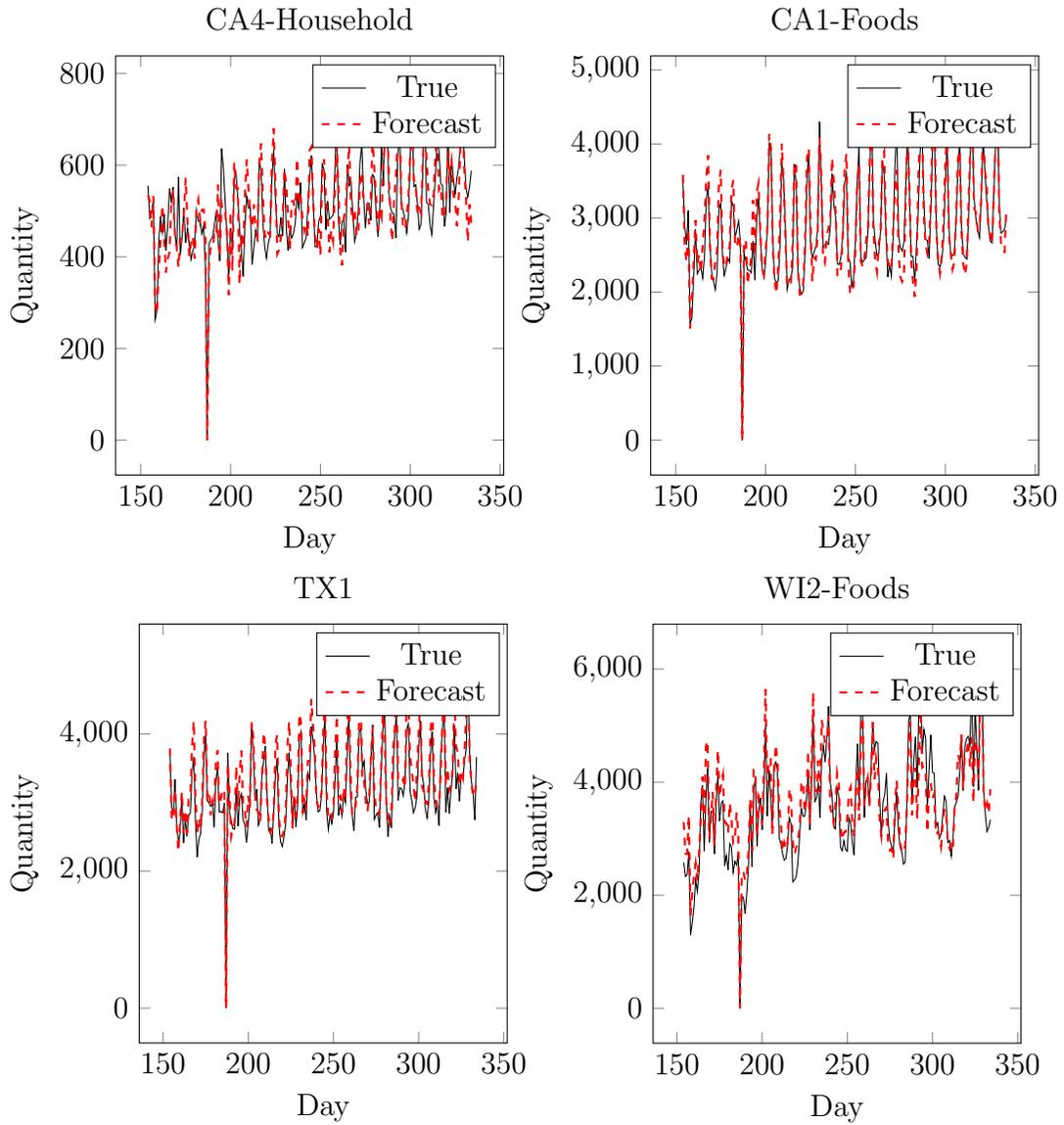

In Figures \ref{fig:italian_plot}, \ref{fig:electricity_plot}, \ref{fig:walmart_plot} we show the NND predictions on the test set for some component series of each dataset.

\begin{figure}

\begin{subfigure}{.5\linewidth}
 
    \begin{tikzpicture}
      \begin{axis}[
      title={M11},
      xlabel={Time},
      ylabel={Power}, 
      width=\textwidth,
      height=\axisdefaultheight]
      
        \addplot[black] table[x=min,y=test_11,col sep=comma]{meter_table_half_week.csv};
        \addplot[thick, red, dashed] table[x=min,y=pred_11,col sep=comma]{meter_table_half_week.csv};
        \legend{True, Forecast}
        
      \end{axis}
      
    \end{tikzpicture}
 
\end{subfigure}%
\begin{subfigure}{.5\linewidth}
 
    \begin{tikzpicture}
      \begin{axis}[
      title={M16},
      xlabel={Time},
      ylabel={Power},
      width=\textwidth,
      height=\axisdefaultheight]
      
        \addplot[black] table[x=min,y=test_16,col sep=comma]{meter_table_half_week.csv};
        \addplot[thick, red, dashed] table[x=min,y=pred_16,col sep=comma]{meter_table_half_week.csv};
        \legend{True, Forecast}
        
      \end{axis}
      
    \end{tikzpicture}
 
\end{subfigure}%

\begin{subfigure}{.5\linewidth}
 
    \begin{tikzpicture}
      \begin{axis}[
      title={M18},
      xlabel={Time},
      ylabel={Power},
      width=\textwidth,
      height=\axisdefaultheight]
      
        \addplot[black] table[x=min,y=test_18,col sep=comma]{meter_table_half_week.csv};
        \addplot[thick, red, dashed] table[x=min,y=pred_18,col sep=comma]{meter_table_half_week.csv};
        \legend{True, Forecast}
        
      \end{axis}
      
    \end{tikzpicture}
 
\end{subfigure}%
\begin{subfigure}{.5\linewidth}
 
    \begin{tikzpicture}
      \begin{axis}[
      title={M19},
      xlabel={Time},
      ylabel={Power},
      width=\textwidth,
      height=\axisdefaultheight]
      
        \addplot[black] table[x=min,y=test_19,col sep=comma]{meter_table_half_week.csv};
        \addplot[thick, red, dashed] table[x=min,y=pred_19,col sep=comma]{meter_table_half_week.csv};
        \legend{True, Forecast}
        
      \end{axis}
      
    \end{tikzpicture}
 
\end{subfigure}
\caption{NND out-of-sample forecasts for the electricity demand dataset. Series M11, M16, M18 and M19 (first 72 hours).}
\label{fig:electricity_plot}
\end{figure}
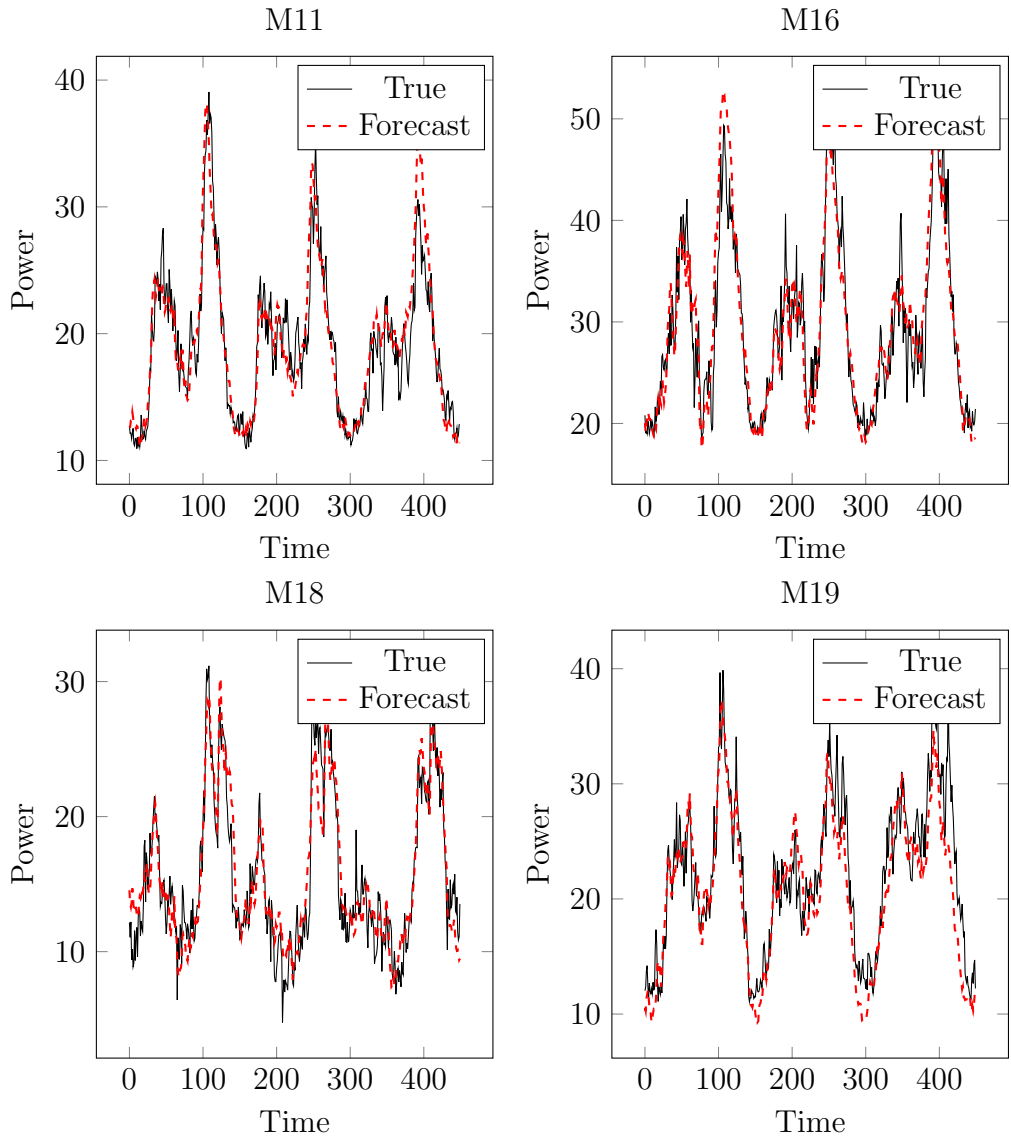

\bibliography{sample}

\end{document}